# Estimating a Function and Its Derivatives Under a Smoothness Condition

**Eunji Lim[a]**

[a] Decision Sciences and Marketing, Adelphi University, Garden City, New York 11530-0701
**Contact:** elim@adelphi.edu, 🆔 https://orcid.org/0000-0003-1008-7050 (EL)

**Abstract.** We consider the problem of estimating an unknown function $f_* : \mathbb{R}^d \to \mathbb{R}$ and its partial derivatives from a noisy data set of $n$ observations, where we make no assumptions about $f_*$ except that it is smooth in the sense that it has square integrable partial derivatives of order $m$. A natural candidate for the estimator of $f_*$ in such a case is the best fit to the data set that satisfies a certain smoothness condition. This estimator can be seen as a least squares estimator subject to an upper bound on some measure of smoothness. Another useful estimator is the one that minimizes the degree of smoothness subject to an upper bound on the average of squared errors. We prove that these two estimators are computable as solutions to quadratic programs, establish the consistency of these estimators and their partial derivatives, and study the convergence rate as $n \to \infty$. The effectiveness of the estimators is illustrated numerically in a setting where the value of a stock option and its second derivative are estimated as functions of the underlying stock price.

## 1. Introduction

We are concerned with the problem of estimating an unknown function $f_* : \mathbb{R}^d \to \mathbb{R}$ and its partial derivatives over a domain $[a,b]^d$ of interest when there is no closed-form formula for $f_*$; hence, simulation must be conducted to estimate $f_*(\mathbf{x})$ at $\mathbf{x} \in \mathbb{R}^d$, or only noisy data on $f_*$ are available. We make no assumptions about $f_*$ except that it has square integrable partial derivatives of order $m$.

To place the problem in a more rigorous setting, we consider the situation where we wish to estimate the unknown function $f_* : \mathbb{R}^d \to \mathbb{R}$ and its partial derivatives by taking observations $(X_1, Y_1), \cdots, (X_n, Y_n)$ satisfying

$$Y_i = f_*(X_i) + \varepsilon_i$$

for $i = 1, 2, \cdots, n$, where $((X_i, \varepsilon_i) : 1 \le i \le n)$ is a sequence of $[a,b]^d \times \mathbb{R}$-valued independent and identically distributed (iid) random vectors satisfying $\mathbb{E}(\varepsilon_i | X_i) = 0$ and $\mathbb{E}(\varepsilon_i^2 | X_i) = \sigma^2 < \infty$ for $i = 1, 2, \cdots, n$.

Under the assumption that $f_*$ has square integrable partial derivatives of order $m$ with $m \ge 1$, our goal is to estimate $f_*$ and its partial derivatives of order up to $m - 1$ as functions over $[a,b]^d$ from the data set $(X_1, Y_1), \cdots, (X_n, Y_n)$.

When the only information available on the unknown regression function $f_*$ is the fact that it is "smooth" so that it has square integrable partial derivatives of order $m$, one way to estimate $f_*$ is to fit a "smooth" function to the data set, that is, find the solution to the following problem:

$$\text{Minimize } E_n(f) \triangleq \frac{1}{n} \sum_{i=1}^{n} (Y_i - f(X_i))^2 \tag{1}$$

over $f \in \mathcal{F}$ for some appropriately chosen set $\mathcal{F}$ of functions. The next question is "Which set is appropriate for $\mathcal{F}$?" One natural candidate for $\mathcal{F}$ is the set $\mathcal{F}'$ of functions $f : \mathbb{R}^d \to \mathbb{R}$ whose partial derivatives of order $m$ are square integrable, that is, $\mathcal{F}' = \{f : \mathbb{R}^d \to \mathbb{R} : J(f) < \infty\}$, where

$$J(f) \triangleq \sum_{|\alpha|=m} \int_{\mathbb{R}^d} \{D^\alpha f(\mathbf{x})\}^2 d\mathbf{x},$$

$\alpha = (\alpha_1, \cdots, \alpha_d)$ is a $d$-tuple of nonnegative integers, $|\alpha|$ is the order of $\alpha$ defined by $|\alpha| = \alpha_1 + \cdots + \alpha_d$, and $D^\alpha$ denotes the partial differential operator defined by

$$D^\alpha = \frac{\partial^{\alpha_1 + \cdots + \alpha_d}}{\partial x_1^{\alpha_1} \cdots \partial x_d^{\alpha_d}}.$$

However, a question arises as to whether there exists a solution to (1). To guarantee the existence of a solution to (1), one of the properties that $\mathcal{F}$ must have is the completeness with respect to a suitably chosen metric (or a pseudometric). It turns out that $\mathcal{F}$ must be larger than $\mathcal{F}'$ to be complete. However, if we let $\mathcal{F}$ be the set of "generalized functions" whose "weak" derivatives of order $m$ are square integrable, then $\mathcal{F}$ contains $\mathcal{F}'$ and is large enough to be a semi-Hilbert space in the $\|\cdot\|_m$ seminorm, where $\|f\|_m \triangleq \sqrt{J(f)}$; see, for example, Meinguet [51, theorem 1 on p. 130]. (A weak derivative of a function $f$ is a generalized version of the classical derivative in the sense that whenever the classical derivative $D^\alpha f$ exists, it is also a weak derivative of $f$.) Hence, throughout this paper, we will assume $\mathcal{F} = \mathcal{F}_m$, where $\mathcal{F}_m$ is the space of generalized functions whose weak partial derivatives of order $m$ are square integrable; see Section 2 of this paper and Oden and Reddy [54] for the precise definitions of $\mathcal{F}_m$, generalized functions, and weak derivatives. We will further restrict our attention to the functions $f$ whose smoothness $J(f)$ is bounded by a certain constant. Thus, a natural candidate for the estimator of $f_*$ is the solution $\hat{f}_n$ to the following problem:

$$\text{Problem (A):} \quad \text{Minimize}_{f \in \mathcal{F}_m} \ E_n(f) \quad \text{subject to } J(f) \leq U_n$$

for some upper bound $U_n$ on $J(f)$. As Proposition 1 of this paper suggests, the solution to Problem (A) exists under the assumption that $2m > d$ and $\{X_1, \cdots, X_n\}$ are $\mathcal{P}_{m-1}$-unisolvent. For definitions, see Assumption 1 in Section 3.

The question now becomes how to compute the solution to Problem (A) numerically. As Proposition 6 of this paper states, the solution to Problem (A) can be found by solving a convex programming problem. However, it is not clear how to compute or estimate $U_n$ from the data set. This necessitates another approach that can estimate $f_*$ using a parameter that can be directly estimated from the data set. As an alternative to Problem (A), we consider the solution $\hat{g}_n$ to the following problem:

$$\text{Problem (B):} \quad \text{Minimize}_{f \in \mathcal{F}_m} \ J(f) \quad \text{subject to } E_n(f) \leq S_n$$

for some upper bound $S_n$ on the average of squared errors. Problem (B) is appealing from a computational point of view; it is intuitively acceptable that $S_n$ should be close to $\sigma^2$, and $\sigma^2$ can be readily estimated from the data set.

In the context of nonparametric regression techniques in the statistics literature, Problems (A) and (B) are often contrasted with the following formulation:

$$\text{Problem (C):} \quad \text{Minimize}_{f \in \mathcal{F}_m} \ E_n(f) + \lambda J(f),$$

where $\lambda$ is a constant called the smoothing parameter. The solution to Problem (C) is often referred to as the "penalized least squares estimator"; see Györfi et al. [27], Wahba [71], and the references therein. One weakness of the penalized least squares estimator is that its performance is highly sensitive to the choice of $\lambda$, and the question of how to choose the smoothing parameter $\lambda$ has been a challenging one. Cross validation (Craven and Wahba [15]) has been a popular method for choosing the smoothing parameter, but it does not guarantee the consistency of the estimators of the derivatives as $n \to \infty$. Thus, when the primary purpose of a modeler is to estimate a derivative of $f_*$ and he or she does not want to use any parameter that cannot be directly estimated from the data set, Problem (B) is preferred from the computational point of view.

This paper is motivated by a stock trader whose goal is to estimate "gamma," which is the second derivative of the value, $f_*$, of a stock option as a function of the underlying stock price. The only information that he has is the fact that gamma is a smooth function of the stock price. The trader can use Problem (C) to estimate $f_*$ and its second derivative, but he needs to estimate the smoothing parameter $\lambda$. Cross validation does not guarantee the consistency of the estimator of the second derivative of $f_*$ as $n \to \infty$. Moreover, the trader does not want to use any parameter that cannot be directly estimated from a data set. In this situation, Problem (B) better serves his purpose because its only parameter $S_n$ can be directly estimated from the data set. In fact, Problem (B) has been used widely within the numerical analysis community; see, for example, Cox [13, p. 530].

Despite its popularity in the numerical analysis community, little is known about how to compute $\hat{f}_n$ and $\hat{g}_n$ numerically and what statistical properties $\hat{f}_n$ and $\hat{g}_n$ possess. The contributions of this paper are showing that $\hat{f}_n$ and $\hat{g}_n$ are computable as solutions to convex programs, establishing the consistency of $\hat{f}_n$ and $\hat{g}_n$ as estimators of $f_*$, establishing the consistency of $D^\alpha \hat{f}_n$ and $D^\alpha \hat{g}_n$ as estimators of $D^\alpha f_*$ for $|\alpha| = 1, \cdots, m-1$, and computing the convergence rate of $\hat{f}_n$ when $n \to \infty$. The key contributions of this paper can be summarized as follows.

1. We identify the relationship between Problems (A) and (B). We prove that for $U_n$ within a certain interval, the solution $\hat{f}_n$ to Problem (A) exists uniquely and becomes a unique solution to Problem (B) with $S_n = J(\hat{f}_n)$. Conversely, for $S_n$ within a certain interval, the solution $\hat{g}_n$ to Problem (B) exists uniquely and becomes a unique solution to Problem (A) with $U_n = E_n(\hat{g}_n)$.

2. We discuss the computational aspects of Problems (A) and (B). We show that the solutions to Problems (A) and (B) can be found by solving convex programming problems. There exist numerous efficient algorithms that can successfully find the solutions to convex programs (see, for example, Boyd and Vandenberghe [6], Zangwill [76]), so this enables us to easily compute the solutions to Problems (A) and (B).

3. We establish the consistency of $\hat{f}_n$ and $\hat{g}_n$ as estimators of $f_*$ and the consistency of $D^\alpha \hat{f}_n$ and $D^\alpha \hat{g}_n$ as estimators of $D^\alpha f_*$ for $|\alpha| = 1, \cdots, m-1$ as $n \to \infty$. Our main results for Problem (A) state that if $U_n \geq J(f_*)$ for $n$ sufficiently large and $\limsup_{n\to\infty} U_n < \infty$, we have

$$\sup_{\mathbf{x} \in [a,b]^d} |\hat{f}_n(\mathbf{x}) - f_*(\mathbf{x})| \to 0 \quad \text{almost surely (a.s.)},$$

and we have

$$\mathbb{E}(\hat{f}_n(X) - f_*(X))^2 \to 0 \text{ and } \mathbb{E}(D^\alpha \hat{f}_n(X) - D^\alpha f_*(X))^2 \to 0$$

for $|\alpha| = 1, \cdots, m-1$ as $n \to \infty$ under modest assumptions. On the other hand, our main results for Problem (B) state that under some conditions on $\hat{g}_n$ and $S_n$, we have

$$\sup_{\mathbf{x} \in [a,b]^d} |\hat{g}_n(\mathbf{x}) - f_*(\mathbf{x})| \to 0 \quad \text{a.s.},$$

$$\mathbb{E}(\hat{g}_n(X) - f_*(X))^2 \to 0 \quad \text{and} \quad \mathbb{E}(D^\alpha \hat{g}_n(X) - D^\alpha f_*(X))^2 \to 0$$

for $|\alpha| = 1, \cdots, m-1$ as $n \to \infty$.

4. We compute the convergence rate of $\hat{f}_n$ as $n \to \infty$. Specifically, when $U_n \geq J(f_*)$ for $n$ sufficiently large and $\limsup_{n\to\infty} U_n < \infty$, we have

$$\left\{ \frac{1}{n} \sum_{i=1}^{n} (\hat{f}_n(X_i) - f_*(X_i))^2 \right\}^{1/2} = \mathcal{O}_p(n^{-m/(2m+d)})$$

under the assumption that the $\varepsilon_i$'s are uniformly sub-Gaussian.

## 1.1. Literature Review

This paper is a contribution to the literature of nonparametric regression for smooth functions. Problem (C) has been studied extensively in the literature, whereas relatively less attention has been paid to Problems (A) and (B). For Problem (C), the existence of a solution is established by Duchon [19], and the convergence rates are derived by Cox [14], Rice and Rosenblatt [57], Utreras [67], and van de Geer [69]. Craven and Wahba [15] and Utreras [66] discuss how to choose the smoothing parameter through cross validation. Duchon [19] has shown that the solution to Problem (C) has an explicit representation as a sum of basis functions. However, this approach often requires solving a system of linear equations that involves a dense and highly ill-conditioned matrix (see Dierckx [17, p. 145]). Numerically more efficient algorithms are developed by Hutchinson and de Hoog [33] and Luo and Wahba [47]. Schoenberg [60] has shown that a solution to Problem (A) is also a solution to Problem (C) for some $\lambda$ appropriately chosen, whereas Kersey [35] and Reinsch [56] studied the relationships between Problems (C) and (B). Green et al. [24] computed convergence rates of the Laplacian smoothing estimator, which can be viewed as a discrete approximation of Problem (C). For further references, see, for example, de Boor [16], Dierckx [17], Green and Silverman [23], Györfi et al. [27], Wahba [71], and Wegman and Wright [74]. For Problem (A), when $d = 1$, the existence and uniqueness of the solution are established by Schoenberg [60], and the convergence rates are derived by van de Geer [68]. When $d \geq 2$, we could not find any work concerning how to compute the solution to Problem (A) numerically or what statistical properties the solution to Problem (A) possesses. For Problem (B), when $d = 1$, the existence and uniqueness of the solution are established by Reinsch [56]. However, we could not find any work concerning how to compute the solution to Problem (B) and what statistical properties the solution to Problem (B) possesses either when $d = 1$ or when $d \geq 2$. Therefore, the main contribution of this paper is to establish a theoretical foundation of Problems (A) and (B) for the case when $d \geq 1$ by suggesting a convex programming formulation, establishing consistency, and computing the convergence rate.

This paper is also a contribution to the literature on shape constrained estimation. Many researchers have studied the problem of estimating the unknown regression function $f_*$ when $f_*$ is known to possess shape properties, such as monotonicity or convexity. One of the most popular approaches is least squares estimation under shape restrictions. For instance, when $f_*$ is known to be monotone, one can fit a monotone function to the data set by solving the following quadratic program:

$$\text{Minimize}_f \; E_n(f) \qquad \text{subject to } f(X_i) \leq f(X_j) \text{ if } X_i \leq X_j \text{ for } 1 \leq i, j \leq n, \tag{2}$$

where $X_i \leq X_j$ denotes coordinate-wise monotonicity. The solution to (2) is referred to as the isotonic regression estimator and has well-established theory. Brunk [7] and Brunk [8] introduced isotonic regression, established consistency, and computed convergence rates when $d = 1$. Mammen [48] considered estimation of a smooth monotone function when $d = 1$. Lim [44] established consistency of isotonic regression and computed its convergence rate when $d \geq 1$ and $f_*$ is possibly misspecified. On the other hand, when $f_*$ is known to be convex, the constraint of (2) can be replaced by the convexity condition, and the solution to the corresponding quadratic program is referred to as the convex regression estimator. Hildreth [30] first introduced convex regression when $d = 1$; Hanson and Pledger [28] established consistency when $d = 1$; Groeneboom et al. [26] computed the rate of convergence when $d = 1$; Kuosmanen [40] formulated the convex regression estimator as the solution to a quadratic program when $d \geq 1$; Lim and Glynn [46] and Seijo and Sen [61] established consistency of the multivariate convex regression estimator; Lim [44] established consistency of convex regression and computed its convergence rates when $d \geq 1$ and $f_*$ is possibly misspecified; Bertsimas and Mundru [4] and Lee et al. [43] proposed computationally efficient algorithms for estimating a concave monotone function and a convex function, respectively; Lim [44] and Yagi et al. [75] proposed hypothesis tests for detecting monotonicity and convexity; Kuosmanen and Johnson [42] discussed an interesting application of production function estimation and proposed a shape constrained estimator for such an application; and Kuosmanen and Johnson [41] established connection between data envelopment analysis and least squares estimation under shape restrictions. Two drawbacks of the convex regression estimator are that (1) it becomes computationally expensive when $n$ increases and that (2) it tends to overfit the data near the boundary of the domain; see, for example, Lim [45, p. 70]. To overcome these shortcomings, Yagi et al. [75] proposed a local polynomial kernel estimator with shape constraints, and Keshvari [36] fitted a convex piecewise linear function to data where the number of linear segments is prespecified. An advantage of Problem (B) is that one can readily compute an estimate of $S_n$ using sample variances. In the context of optimizing behavior in economics, Varian [70] similarly used sample variances to set bounds for acceptable deviations from the fitted function. For a comprehensive survey, see Johnson and Jiang [34] and the references therein. Although the previous literature has focused on monotonicity and convexity constraints, this paper demonstrates that one can work with the smoothness condition only. In particular, Problem (A) replaces the constraint of (2) by a condition on the degree of smoothness (i.e., $J(f) \leq U_n$). Therefore, this paper can be viewed as an addition to the literature on nonparametric regression subject to shape constraints.

This paper can be viewed as a contribution to the literature on metamodel estimation or response surface estimation, which has been an important subject of research in the simulation community. A common goal in this setting is to fit a metamodel to a simulation output, where the simulation runs are time consuming. Therefore, most work has focused on a setting where $n$ is relatively small. One of the earlier works is the response surface methodology that is based on the idea of fitting a polynomial function to the data set locally; see, for example, Myers and Montgomery [52]. When the $Y_i$'s are possibly correlated, the kriging-based methods serve as good alternatives; see, for example, Ankenman et al. [3]. The kriging-based methods, in general, do not assume that the regression function $f_*$ is $m$ times differentiable, where $m \geq 2$. Hence, they do not produce any estimator of the derivatives of $f_*$. Rather, the kriging estimator of $f_*(\mathbf{x})$ at any $\mathbf{x} \in [a, b]^d$ is expressed as the weighted sum of the $Y_i$'s. The weights are dependent on $\mathbf{x}_0$ and calculated in a way in which the mean squared deviations are minimized. In Chen et al. [9], under the assumption that the regression function is differentiable, the gradient estimates are observed along with the $Y_i$'s, and they are used to obtain an estimator of $f_*$ that shows superior performance to the stochastic kriging estimator (see Ankenman et al. [3]). In the radial basis function estimators, the regression function is approximated by a linear combination of radially symmetric functions, where the coefficients and other parameters are estimated from the data set; see, for example, Franke [21]. The orthogonal series estimators or wavelet estimators represent the regression function by its Fourier series and estimate the coefficients from the data set; see, for example, Donoho and Johnstone [18]. Nonparametric regression methods, such as the kernel regression method (Nadaraya [53], Watson [73]), the local polynomial regression estimators (Cleveland [12]), the smoothing splines (Reinsch [55]), the weighted least squares regression estimators (Ruppert and Wand [58], Salemi et al. [59]), and the nearest neighbor estimators (Shapiro [62], Stone [64]), have received considerable attention in the statistics literature; see, for example, Eubank [20], Green and Silverman [23], Härdle [29], and Wand and Jones [72] for comprehensive surveys. Most of these estimators invariably require the estimation of smoothing parameters that cannot be estimated directly from the data set. In this regard, Problem (B) has the advantage of estimating only one parameter, $S_n$, which has a direct meaning and can be estimated from the sample variances computed from the data set.

### 1.2. Organization of This Paper

This paper is organized as follows. Section 2 introduces some definitions and preliminaries. We precisely describe the relationship between Problems (A) and (B) in Section 3, whereas Section 4 is concerned with the convex programming formulations of Problems (A) and (B). Section 5 states the main results regarding consistency and convergence rates, and the numerical results can be found in Section 6. Section 7 provides discussions on future research topics. The proofs of all theorems, corollaries, lemmas, and propositions are provided in the Appendix.

## 2. Definitions and Preliminaries

For $\mathbf{x} = (x_1, \cdots, x_d) \in \mathbb{R}^d$, its norm is given by $\|\mathbf{x}\| = \{\sum_{j=1}^n x_j^2\}^{1/2}$. For a sequence of random variables $(Z_n : n \geq 1)$ and a sequence of positive real numbers $(\alpha_n : n \geq 1)$, we say $Z_n = \mathcal{O}_p(\alpha_n)$ as $n \to \infty$ if for any $\epsilon > 0$, there exist constants $C$ and $N$ such that $\mathbb{P}(|Z_n/\alpha_n| > C) < \epsilon$ for all $n \geq N$.

When $2m > d$, we define $\mathcal{F}_m$ by the space of generalized functions whose weak partial derivatives of order $m$ are square integrable, that is,

$$\mathcal{F}_m = \left\{ f \in \mathcal{D}'(\mathbb{R}^d) : \int_{\mathbb{R}^d} \{D^\alpha f(\mathbf{x})\}^2 d\mathbf{x} < \infty \text{ for all } \alpha \text{ such that } |\alpha| = m \right\},$$

where $\mathcal{D}'(\mathbb{R}^d)$ is the space of Schwartz distributions. A test function on $\mathbb{R}^d$ is a real-valued, infinitely differentiable function with compact support. Let $\mathcal{D}(\mathbb{R}^d)$ be the set of all test functions on $\mathbb{R}^d$. A functional on $\mathcal{D}(\mathbb{R}^d)$ is a mapping that assigns to each $\phi \in \mathcal{D}(\mathbb{R}^d)$ a real number. A (Schwartz) distribution on $\mathbb{R}^d$ is a continuous linear functional on $\mathcal{D}(\mathbb{R}^d)$. See Adams and Fournier [2, p. 20] or Oden and Reddy [54, p. 14] for detailed definitions. Let $(\cdot, \cdot)_m$ denote the semi-inner product on $\mathcal{F}_m$, defined by

$$(f, g)_m \triangleq \sum_{|\alpha| = m} \int_{\mathbb{R}^d} \{D^\alpha f(\mathbf{x})\}\{D^\alpha g(\mathbf{x})\} d\mathbf{x}$$

for $f, g \in \mathcal{F}_m$. The associated seminorm $\|\cdot\|_m$ is defined by $\|f\|_m \triangleq (f, f)_m^{1/2}$. It should be noted that the kernel of this seminorm is the vector space $\mathcal{P}_{m-1}$ of all polynomials of total degree less than or equal to $m - 1$ defined on $\mathbb{R}^d$, the dimension of $\mathcal{P}_{m-1}$ is

$$M \triangleq \binom{m + d - 1}{d}, \tag{3}$$

and $\mathcal{P}_{m-1}$ is spanned by the $M$ monomials of the total degree less than or equal to $m - 1$. We will denote the $M$ monomials of total degree less than or equal to $m - 1$ by $p_1, \cdots, p_M$.

It should be also noted that $\|\cdot\|_m$ is a seminorm on $\mathcal{F}_m$, but a norm on the equivalence classes in $\mathcal{F}_m/\mathcal{P}_{m-1}$, and $\mathcal{F}_m/\mathcal{P}_{m-1}$ is a Hilbert space under $(\cdot, \cdot)_m$; see Meinguet [51, theorem 1, p. 130] for details.

For any open-bounded subset $\Omega \subset \mathbb{R}^d$, we define $L_2(\Omega)$ by the set of functions $f$ defined on $\Omega$ satisfying $\int_\Omega \{f(\mathbf{x})\}^2 d\mathbf{x} < \infty$. Its $L_2(\Omega)$ norm is defined as follows:

$$\|f\|_{L_2(\Omega)} = \left( \int_\Omega \{f(\mathbf{x})\}^2 d\mathbf{x} \right)^{1/2}$$

for $f \in L_2(\Omega)$. We also define $W_m(\Omega)$ by the set of functions whose weak partial derivatives up to order $m$ are in $L_2(\Omega)$, that is,

$$W_m(\Omega) = \{f \in L_2(\Omega) : D^\alpha f \in L_2(\Omega) \text{ for all } \alpha \text{ such that } |\alpha| \leq m\}.$$

Its $W_m(\Omega)$ norm is defined as follows:

$$\|f\|_{W_m(\Omega)} = \left( \int_\Omega \sum_{|\alpha| \leq m} \{D^\alpha f(\mathbf{x})\}^2 d\mathbf{x} \right)^{1/2}$$

for $f \in W_m(\Omega)$. $W_m(\Omega)$ is referred to as a Sobolev space of order $m$ on $\Omega$.

## 3. Relationships Between Problems (A) and (B)

In this section, we study the relationship between Problems (A) and (B). When $d = 1$, Schoenberg [60] established some results similar to Lemma 1, Proposition 3, Proposition 4, and Proposition 5 of this paper. Thus, some of our results in this section can be viewed as extensions of those in Schoenberg [60] to the multidimensional case. Throughout this section, we assume that $(X_1, Y_1), \cdots, (X_n, Y_n)$ are given and that the following assumption holds.

**Assumption 1.** *$m$ is a positive integer satisfying $2m > d$, and $\{X_1, \cdots, X_n\}$ is a set of mutually distinct points containing a $\mathcal{P}_{m-1}$-unisolvent set. A set of points $\{\mathbf{x}_1, \cdots, \mathbf{x}_L\} \subset \mathbb{R}^d$ is called $\mathcal{P}_{m-1}$-unisolvent if for any $p \in \mathcal{P}_{m-1}$, the condition $p(\mathbf{x}_j) = 0$ for all $1 \le j \le L$ implies $p(\mathbf{x}) = 0$ for all $\mathbf{x} \in \mathbb{R}^d$. In particular, when $d = 1$, $\{\mathbf{x}_1, \cdots, \mathbf{x}_L\}$ is $\mathcal{P}_{m-1}$-unisolvent if $L \ge m$ and $\mathbf{x}_1, \cdots, \mathbf{x}_L$ are mutually distinct.*

Propositions 1 and 2 establish the existence of the solutions to Problems (A) and (B) under Assumption 1.

**Proposition 1.** *Let $U_n \ge 0$ be given. Under Assumption 1, there exists a solution to Problem (A). Furthermore, if $\hat{f}_n$ and $\tilde{f}_n$ are two solutions to Problem (A), then $\hat{f}_n(X_i) = \tilde{f}_n(X_i)$ for $1 \le i \le n$.*

**Proposition 2.** *Let $S_n \ge 0$ be given. Under Assumption 1, there exists a solution to Problem (B). Furthermore, if $\hat{g}_n$ and $\tilde{g}_n$ are two solutions to Problem (B), then they differ by a polynomial of total degree less than or equal to $m - 1$ (i.e., $\hat{g}_n - \tilde{g}_n \in \mathcal{P}_{m-1}$).*

Our next propositions, Propositions 3, 4, and 5, are concerned with the uniqueness of the solutions to Problems (A) and (B). For our analysis, we need to define the following function: $\Psi_n : [0, \infty) \to \mathbb{R}$. For any nonnegative real number $u$, let $\Psi_n(u) = E_n(f^u)$, where $f^u$ is a solution to Problem (A) with $U_n = u$. The following lemma characterizes the shape of $\Psi_n$.

**Lemma 1.** *There exists a minimizer, say $f_P$, of $E_n(f)$ over $f \in \mathcal{P}_{m-1}$. Under Assumption 1, there exists a unique solution, say $f_I$, to the following problem:*

$$Minimize_{f \in \mathcal{F}_m} J(f) \text{ subject to } f(X_i) = Y_i \quad 1 \le i \le n.$$

*Under Assumption 1, $\Psi_n$ is nonnegative, convex, and strictly decreasing over $[0, J(f_I)]$ with $\Psi_n(0) = E_n(f_P)$ and $\Psi_n(x) = 0$ for all $x \ge J(f_I)$.*

The following propositions discuss the uniqueness of the solutions to Problems (A) and (B) and the relationship between them.

**Proposition 3.** *Assume Assumption 1. Let $0 \le U_n < J(f_I)$ be given. Then, there exists a unique solution $\hat{f}_n$ to Problem (A), and $\hat{f}_n$ satisfies $J(\hat{f}_n) = U_n$. In other words, if $\hat{f}_n$ and $\tilde{f}_n$ are two solutions to Problem (A), then $\hat{f}_n(\mathbf{x}) = \tilde{f}_n(\mathbf{x})$ for all $\mathbf{x} \in \mathbb{R}^d$. Furthermore, $\hat{f}_n$ is also a solution to Problem (B) with $S_n = E_n(\hat{f}_n)$.*

**Proposition 4.** *Assume Assumption 1. Let $0 < S_n \le E_n(f_P)$ be given. Then, there exists a unique solution $\hat{g}_n$ to Problem (B), and $\hat{g}_n$ satisfies $E_n(\hat{g}_n) = S_n$. In other words, if $\hat{g}_n$ and $\tilde{g}_n$ are two solutions to Problem (B), then $\hat{g}_n(\mathbf{x}) = \tilde{g}_n(\mathbf{x})$ for all $\mathbf{x} \in \mathbb{R}^d$. Furthermore, $\hat{g}_n$ is also a solution to Problem (A) with $U_n = J(\hat{g}_n)$.*

Combining Propositions 3 and 4 yields the following proposition.

**Proposition 5.** *Assume Assumption 1.*
i. *Let $0 \le U_n < J(f_I)$ be given. Then, there exists a unique solution $\hat{f}_n$ to Problem (A), and $\hat{f}_n$ satisfies $J(\hat{f}_n) = U_n$. Furthermore, $\hat{f}_n$ is also a unique solution to Problem (B) with $S_n = E_n(\hat{f}_n)$.*
ii. *Let $0 < S_n \le E_n(f_P)$ be given. Then, there exists a unique solution $\hat{g}_n$ to Problem (B), and $\hat{g}_n$ satisfies $E_n(\hat{g}_n) = S_n$. Furthermore, $\hat{g}_n$ is also a unique solution to Problem (A) with $U_n = J(\hat{g}_n)$.*

The proofs of Lemma 1 and Propositions 1–5 are provided in Appendix A.

## 4. Convex Programming Formulations for Problems (A) and (B)

In this section, we discuss the question of how to compute the solutions to Problems (A) and (B) numerically. The following propositions, Propositions 6 and 7, reveal that we can find the solutions to Problems (A) and (B) by solving convex programs. Their proofs are provided in Appendix A.

We start with some definitions. Let $\mathbf{s}_1, \cdots, \mathbf{s}_M$ be any $\mathcal{P}_{m-1}$-unisolvent set of $M$ fixed points in $\mathbb{R}^d$. ($M$ is given in (3).) Let $q_1, \cdots, q_M$ be the unique polynomials of total degree less than $m$ satisfying $q_i(\mathbf{s}_j) = 1$ if $i = j$ and $q_i(\mathbf{s}_j) = 0$ if $i \neq j$, that is,

$$
q_i(\mathbf{x}) = \begin{bmatrix} p_1(\mathbf{x}) & \cdots & p_M(\mathbf{x}) \end{bmatrix} \begin{bmatrix} p_1(\mathbf{s}_1) & \cdots & p_M(\mathbf{s}_1) \\ \vdots & & \vdots \\ p_1(\mathbf{s}_M) & \cdots & p_M(\mathbf{s}_M) \end{bmatrix}^{-1} \cdot e_i
$$

for $\mathbf{x} \in \mathbb{R}^d$ and $1 \leq i \leq M$, where $e_i$ is the $M \times 1$ column vector whose $i$th entry is one and zero elsewhere. Let $R : \mathbb{R}^d \times \mathbb{R}^d \to \mathbb{R}$ be defined by

$$
R(\mathbf{s}, \mathbf{t}) = (-1)^m \left[ K_m(\mathbf{s} - \mathbf{t}) - \sum_{i=1}^{M} q_i(\mathbf{t}) K_m(\mathbf{s} - \mathbf{s}_i) - \sum_{j=1}^{M} q_j(\mathbf{s}) K_m(\mathbf{s}_j - \mathbf{t}) + \sum_{i=1}^{M} \sum_{j=1}^{M} q_i(\mathbf{s}) q_j(\mathbf{t}) K_m(\mathbf{s}_i - \mathbf{s}_j) \right]
$$

for $\mathbf{s}, \mathbf{t} \in \mathbb{R}^d$, where $K_m : \mathbb{R}^d \to \mathbb{R}$ is given by

$$
K_m(\mathbf{z}) = \begin{cases} \theta_{m,d} \|\mathbf{z}\|^{2m-d} \ln \|\mathbf{z}\|, & \text{if } 2m - d \text{ is even} \\ \theta_{m,d} \|\mathbf{z}\|^{2m-d}, & \text{otherwise} \end{cases}
$$

for $\mathbf{z} \in \mathbb{R}^d$,

$$
\theta_{m,d} = \begin{cases} \dfrac{(-1)^{d/2+1}}{2^{2m-1} \pi^{d/2} (m-1)! (m-d/2)!}, & \text{if } 2m - d \text{ is even} \\ \dfrac{(-1)^m \Gamma(d/2 - m)}{2^{2m} \pi^{d/2} (m-1)!}, & \text{otherwise} \end{cases}
$$

and $\Gamma(\cdot)$ denotes the gamma function.

**Proposition 6.** *Let $U_n \geq 0$ be given. Under Assumption* 1, *a solution to Problem* (A) *exists by Proposition* 1. *Furthermore, for any solution $\tilde{f}_n$ to Problem* (A), *there exists a solution $\hat{f}_n$ to Problem* (A), *satisfying $\tilde{f}_n(X_i) = \hat{f}_n(X_i)$ for $1 \leq i \leq n$, which can be represented by*

$$
\hat{f}_n(\mathbf{x}) = \sum_{i=1}^{M} \hat{c}_i p_i(\mathbf{x}) + \sum_{i=1}^{n} \hat{d}_i R(\mathbf{x}, X_i) \tag{4}
$$

*for $\mathbf{x} \in \mathbb{R}^d$, where $(\hat{c}_1, \cdots, \hat{c}_M, \hat{d}_1, \cdots, \hat{d}_n, \hat{f}_n(X_1), \cdots, \hat{f}_n(X_n))$ is a solution to the following convex program in the decision variables $c_1, \cdots, c_M, d_1, \cdots, d_n, y_1, \cdots, y_n \in \mathbb{R}$.*

$$
\text{Minimize } \frac{1}{n} \sum_{i=1}^{n} (Y_i - y_i)^2
$$

$$
\text{subject to } \sum_{i=1}^{n} \sum_{j=1}^{n} R(X_i, X_j) d_i d_j \leq U_n
$$

$$
\sum_{i=1}^{M} c_i p_i(X_j) + \sum_{i=1}^{n} d_i R(X_j, X_i) = y_j \quad 1 \leq j \leq n. \tag{5}
$$

*Conversely, any function of form* (4), *where $(\hat{c}_1, \cdots, \hat{c}_M, \hat{d}_1, \cdots, \hat{d}_n, \hat{f}_n(X_1), \cdots, \hat{f}_n(X_n))$ is any solution to* (6), *is a solution to Problem* (A).

**Proposition 7.** *Let $S_n \geq 0$ be given. Under Assumption* 1, *a solution to Problem* (B) *exists by Proposition* 2. *Furthermore, $\hat{g}_n$ is a solution to Problem* (B) *if and only if $\hat{g}_n$ is represented by*

$$
\hat{g}_n(\mathbf{x}) = \sum_{i=1}^{M} \hat{c}_i p_i(\mathbf{x}) + \sum_{i=1}^{n} \hat{d}_i R(\mathbf{x}, X_i) \tag{6}
$$

for $\mathbf{x} \in \mathbb{R}^d$, where $(\hat{c}_1, \cdots, \hat{c}_M, \hat{d}_1, \cdots, \hat{d}_n, \hat{g}_n(X_1), \cdots, \hat{g}_n(X_n))$ is a solution to the following convex program in the decision variables $c_1, \cdots, c_M, d_1, \cdots, d_n, y_1, \cdots, y_n \in \mathbb{R}$.

$$Minimize \sum_{i=1}^{n}\sum_{j=1}^{n} R(X_i, X_j)d_i d_j$$

$$subject\ to\ \frac{1}{n}\sum_{i=1}^{n}(Y_i - y_i)^2 \leq S_n$$

$$\sum_{i=1}^{M} c_i p_i(X_j) + \sum_{i=1}^{n} d_i R(X_j, X_i) = y_j \quad 1 \leq j \leq n.$$

**Remark 1.** It should be noted that when $d = 1$, the representation in (4) coincides with the form of a natural spline; see Greville [25, equation 4.1 on p. 3] for the details.

**Remark 2.** Proposition 6 reveals that for any solution $\tilde{f}_n$ to Problem (A), there exists a solution $\hat{f}_n$ to Problem (A) satisfying $\tilde{f}_n(X_i) = \hat{f}_n(X_i)$ for $1 \leq i \leq n$, which is a linear combination of the $p_i$'s and $R(\cdot, X_i)$'s. In addition, Proposition 7 implies that a solution to Problem (B) is always a linear combination of the $p_i$'s and $R(\cdot, X_i)$'s.

## 5. Consistency and Rates of Convergence

In this section, we focus on the consistency and convergence rates of the solutions to Problems (A) and (B). Toward this goal, we need the following assumptions.

**Assumption 2.**
i. $2m > d$.
ii. $X, X_1, X_2, \cdots$ are iid $[a,b]^d$-valued random vectors having a common positive continuous density function $\tau : [a,b]^d \to \mathbb{R}$. This implies that there exist $\tau_1, \tau_2 > 0$ such that $\tau_1 \leq \tau(x) \leq \tau_2$ for all $x \in [a,b]^d$.

**Assumption 3.**
i. $f_* \in \mathcal{F}_m$.
ii. $(X,Y),(X_1,Y_1),(X_2,Y_2), \cdots$ is a sequence of iid $[a,b]^d \times \mathbb{R}$-valued random vectors satisfying $Y = f_*(X) + \varepsilon$ and $Y_i = f_*(X_i) + \varepsilon_i$ for $i = 1, 2, \cdots$.
iii. $\mathbb{E}(\varepsilon|X) = \mathbb{E}(\varepsilon_i|X_i) = 0$ and $\mathbb{E}(\varepsilon^2|X) = \mathbb{E}(\varepsilon_i^2|X_i) = \sigma^2 < \infty$ for $i = 1, 2, \cdots$.

**Assumption 4.** $\varepsilon, \varepsilon_1, \varepsilon_2, \cdots$ are uniformly sub-Gaussian random variables (i.e., there exist positive real numbers A and B such that $\mathbb{E}(\exp(t\varepsilon)) \leq A\exp(Bt^2)$ for any $t \in \mathbb{R}$).

**Assumption 5.**
i. $\mathbb{E}(f_*(X)^2) < \infty$.
ii. $f_*$, restricted to $[a,b]^d$, is not a polynomial of degree less than or equal to $m-1$.
iii. $f_*$ is continuous on $[a,b]^d$.

We need Assumption 2(i) for the following reason. When $2m > d$ is not assumed, a function in $\mathcal{F}_m$ is not well defined on a set of measure 0 because the functions in $\mathcal{F}_m$ are determined only outside a set of measure 0. Under the assumption $2m > d$, it is known that the functions in $\mathcal{F}_m$ are continuous, so the point evaluation is well defined; see Meinguet [51, theorem 1 on p. 130].

The next lemma shows that Assumption 2 guarantees a very important property of the $X_i$'s, the so-called $\mathcal{P}_{m-1}$-unisolvency. The proof of Lemma 2 is provided in Appendix A.

**Lemma 2.** Under Assumption 2, $\{X_1, \cdots, X_n\}$ is a set of mutually distinct points containing a $\mathcal{P}_{m-1}$-unisolvent set for $n$ sufficiently large a.s. In other words, Assumption 2 implies Assumption 1 for $n$ sufficiently large a.s.

Lemma 2 ensures that the solutions to Problems (A) and (B) exist for $n$ sufficiently large a.s. under Assumption 2.

We are now ready to present our main results. Theorems 1 and 2 reveal that $\hat{f}_n$ and $\hat{g}_n$ are consistent estimators of $f_*$ and that $D^\alpha \hat{f}$ and $D^\alpha \hat{g}_n$ are consistent estimators of $D^\alpha f_*$ for $|\alpha| = 1, \cdots, m-1$. Theorem 3 computes a lower bound on the rate of convergence of $\hat{f}_n$. We discuss the rate of convergence of $\hat{g}_n$ in Remark 4.

Theorems 1 and 2 use the following assumptions on $\hat{f}_n$ and $\hat{g}_n$.

**Assumption 6.** There exists a constant $c_A$ such that $\lim\sup_{n\to\infty} J(\hat{f}_n) \leq c_A$ for all $n$ sufficiently large a.s.

**Assumption 7.** *There exists a constant $c_B$ such that* $\limsup_{n\to\infty} J(\hat{g}_n) \le c_B$ *for all n sufficiently large a.s.*

**Assumption 8.**

$$\frac{1}{n}\sum_{i=1}^{n}(\hat{f}_n(X_i) - f_*(X_i))^2 \le \frac{2}{n}\sum_{i=1}^{n}\varepsilon_i(\hat{f}_n(X_i) - f_*(X_i)) + A_n$$

*for a sequence of random variables $(A_n : n \ge 1)$ satisfying* $\limsup_{n\to\infty} A_n \le 0$ *a.s.*

**Assumption 9.**

$$\frac{1}{n}\sum_{i=1}^{n}(\hat{g}_n(X_i) - f_*(X_i))^2 \le \frac{2}{n}\sum_{i=1}^{n}\varepsilon_i(\hat{g}_n(X_i) - f_*(X_i)) + B_n$$

*for a sequence of random variables $(B_n : n \ge 1)$ satisfying* $\limsup_{n\to\infty} B_n \le 0$ *a.s.*

Although Assumptions 6–9 are somewhat hard to verify, we can establish them by imposing some conditions on $U_n$ and $S_n$. For example, Assumptions 6, 8, and 9 are implied by Assumptions 10, 11, and 12, respectively.

**Assumption 10.** $\limsup_{n\to\infty} U_n < \infty.$

**Assumption 11.** $J(f_*) \le U_n$ *for all n sufficiently large a.s.*

**Assumption 12.** $\limsup_{n\to\infty} S_n \le \sigma^2.$

We are now ready to present Theorem 1, Theorem 2, Corollary 1, and Corollary 2.

**Theorem 1.** *Under Assumptions 2, 3, 6, and 8,*

$$\sup_{\mathbf{x}\in[a,b]^d} |\hat{f}_n(\mathbf{x}) - f_*(\mathbf{x})| \to 0 \quad a.s., \tag{7}$$

$$\mathbb{E}(\hat{f}_n(X) - f_*(X))^2 \to 0 \quad and \quad \mathbb{E}(D^\alpha\hat{f}_n(X) - D^\alpha f_*(X))^2 \to 0 \tag{8}$$

*for* $|\alpha| = 1, \cdots, m-1$ *as* $n \to \infty$.

**Theorem 2.** *Under Assumptions 2, 3, 7, and 9,*

$$\sup_{\mathbf{x}\in[a,b]^d} |\hat{g}_n(\mathbf{x}) - f_*(\mathbf{x})| \to 0 \quad a.s., \tag{9}$$

$$\mathbb{E}(\hat{g}_n(X) - f_*(X))^2 \to 0 \quad and \quad \mathbb{E}(D^\alpha\hat{g}_n(X) - D^\alpha f_*(X))^2 \to 0 \tag{10}$$

*for* $|\alpha| = 1, \cdots, m-1$ *as* $n \to \infty$.

**Corollary 1.** *Under Assumptions 2, 3, 10, and 11, (7) and (8) hold.*

**Corollary 2.** *Under Assumptions 2, 3, 7, and 12, (9) and (10) hold.*

We now turn to the convergence rate of $\hat{f}_n$. We need Assumption 13, which is implied by Assumption 11.

**Assumption 13.**

$$\frac{1}{n}\sum_{i=1}^{n}(\hat{f}_n(X_i) - f_*(X_i))^2 \le \frac{2}{n}\sum_{i=1}^{n}\varepsilon_i(\hat{f}_n(X_i) - f_*(X_i))$$

*for all n sufficiently large a.s.*

Under Assumption 13, the convergence rate of $\hat{f}_n$ is established in Theorem 3.

**Theorem 3.** *Under Assumptions 2, 3, 4, 6, and 13,*

$$\left\{\frac{1}{n}\sum_{i=1}^{n}(\hat{f}_n(X_i) - f_*(X_i))^2\right\}^{1/2} = \mathcal{O}_p(n^{-m/(2m+d)}) \tag{11}$$

*as* $n \to \infty$.

**Corollary 3.** *Under Assumptions 2, 3, 4, 10, and 11, (11) holds.*

**Remark 3.** It should be noted that the rate $-m/(2m + d)$ established in Theorem 3 is identical to the optimal rate of convergence for the estimation of a function of smoothness of order $m$, which was established by Stone [65]. However, as Corollary 3 states, this rate is achieved under Assumptions 10 and 11, which may be hard to verify if sufficient information on $J(f_*)$ is not available. One should also note that there are many estimators that achieve this optimal rate. See Kohler et al. [38] for a kernel estimator; Cox [14] for a smoothing spline estimator; Györfi et al. [27, theorems 4.3, 5.2., and 6.2] for partitioning, kernel, and nearest neighbor estimators, respectively; and Kohler et al. [37] for partitioning and nearest neighbor estimators. Furthermore, Stone [65] established that the optimal rate of convergence for the estimation of the $r$th derivative of a function of smoothness of order $m$ is $-(m - r)/(2m + d)$ for $r = 0, 1, \cdots, m - 1$, and this rate is achieved by the smoothing spline estimator proposed by Cox [14].

We next discuss a situation where the solution to Problem (B) exists uniquely.

**Theorem 4.** *Consider the problem:*

$$\text{Minimize } \mathbb{E}(c_1 p_1(X) + \cdots + c_M p_M(X) - f_*(X))^2 \tag{12}$$

*over $c_1, \cdots, c_M \in \mathbb{R}$. Under Assumption 5, (12) has a solution, and the minimizing value $v^*$ of (12) is positive. Furthermore, if $\limsup_{n\to\infty} S_n < \sigma^2 + v^*$ a.s., under Assumption 2, Assumption 3(ii), Assumption 3(iii), and Assumption 5, the solution to Problem (B) exists uniquely for $n$ sufficiently large a.s.*

The proofs of Theorems 1–4 and Corollaries 1–3 are provided in Appendix B.

**Remark 4.** The same convergence rate as in Theorem 3 can be obtained for $\hat{g}_n$ under the assumption

$$\frac{1}{n}\sum_{i=1}^{n}(\hat{g}_n(X_i) - f_*(X_i))^2 \leq \frac{2}{n}\sum_{i=1}^{n}\varepsilon_i(\hat{g}_n(X_i) - f_*(X_i)) \tag{13}$$

for all $n$ sufficiently large a.s. (Arguments similar to those in the proof of Theorem 3 can be applied.) However, we were not able to find a suitable condition on $S_n$ under which (13) holds.

**Remark 5.** In the case where the error $\varepsilon_1$ is dependent on $X_1$, our results in Theorems 1, 2, 3, and 4 hold with light modifications in the proofs.

**Remark 6.** In many practical situations, the variance of the $Y_i$'s is heterogeneous (i.e., $\text{Var}(\varepsilon_1)$ is dependent on $X_1$). A natural candidate for $S_n$ in such a case is $S_n' \triangleq \mathbb{E}(\text{Var}(\varepsilon_1 | X_1)) = \mathbb{E}(\text{Var}(Y_1 | X_1)) = \int_{[a,b]^d} \text{Var}(Y_1 | \mathbf{x})\tau(\mathbf{x})d\mathbf{x}$, where $\tau(\cdot)$ is the density function of $X_1$. When the $X_i$'s are uniformly distributed, $S_n'$ can be estimated by computing the average of $\text{Var}(Y_i | X_i)$ across $i \in \{1, \cdots, n\}$.

## 6. Numerical Results

In this section, we observe the numerical behavior of the solutions to Problems (A), (B), and (C). To describe the detailed procedure, we start by noticing that, in practice, we can often observe $Y = f_*(X) + \varepsilon$ multiple times at a fixed design point $X \in [a, b]^d$. Thus, we assume that we can collect iid observations $((X_i, Y_{ij}) : 1 \leq i \leq n, 1 \leq j \leq r)$, where $Y_{ij} = f_*(X_i) + \varepsilon_{ij}$ for $1 \leq i \leq n$ and $1 \leq j \leq r$ and the $\varepsilon_{ij}$'s, conditional on $X_i$, are iid copies of $\varepsilon$ conditional on $X$. We then compute the average of the $Y_{ij}$'s across multiple replications (i.e., $\overline{Y}_i = \sum_{j=1}^{r} Y_{ij}/r$ for each $i \in \{1, \cdots, n\}$). Next, the $\overline{Y}_i$'s will be used in place of the $Y_i$'s in Problems (A), (B), and (C).

Next, we provide more detailed descriptions on how the solutions to Problems (A)–(C) are obtained.

### 6.1. Problem (B)

In order to estimate $S_n$, we observe that Remark 6 suggests using the average of the $\text{Var}(Y_i | X_i)$'s as an estimate of $S_n$ when the $X_i$'s are uniformly distributed. Because we use the $\overline{Y}_i$'s in place of the $Y_i$'s, we use the average of $\text{Var}(\overline{Y}_i | X_i) = (1/r)\text{Var}(Y_i | X_i)$ across $i \in \{1, \cdots, n\}$ as an estimate of $S_n$. Therefore, our estimate of $S_n$ is

$$\frac{1}{nr}\sum_{i=1}^{n}\hat{S}_i, \tag{14}$$

where $\hat{S}_i = \sum_{j=1}^{r}(Y_{ij} - \overline{Y}_i)^2/(r - 1)$.

We then solve the convex programming problem in Proposition 7 with the $Y_i$'s replaced by the $\overline{Y}_i$'s and $S_n$ replaced by (14) using CVX, a package for solving convex programs developed by Grant and Boyd [22]. The solution obtained this way is denoted by $\hat{g}_n$, and its objective value $J(\hat{g}_n)$ is denoted by $\hat{U}_n$. $\hat{U}_n$ is used as an estimate of $U_n$ when solving Problem (A).

## 6.2. Problem (A)

We next solve the convex programming problem in Proposition 6 by using $\hat{U}_n = J(\hat{g}_n)$ as an estimate of $U_n$ and the $\overline{Y}_i$'s in place of the $Y_i$'s. CVX is used to solve the convex programming formulation.

## 6.3. Problem (C)

The key issue here is how to determine the smoothing parameter $\lambda$. We use cross validation to determine the smoothing parameter. We also use a fixed sequence of real numbers $(\lambda_n : n \geq 1)$ decreasing to zero as $n \to \infty$. The smoothing parameter $\lambda$ chosen by cross validation depends on the observed data set $((X_i, Y_{ij}) : 1 \leq i \leq n, 1 \leq j \leq r)$, whereas $\lambda_n$ does not depend on the observed data set and depends on $n$ only.

To define the cross validation function, let $f_\lambda^{[k]}$ be the minimizer of

$$\frac{1}{n} \sum_{i=1, i \neq k}^{n} (\overline{Y}_i - f(X_i))^2 + \lambda J(f)$$

over $f \in \mathcal{F}_m$. Then, the cross validation function $V(\lambda)$ is defined by

$$V(\lambda) = \frac{1}{n} \sum_{k=1}^{n} (\overline{Y}_k - f_\lambda^{[k]}(X_k))^2 \tag{15}$$

for $\lambda \geq 0$, and the estimator of the smoothing parameter based on cross validation is the minimizer of $V(\lambda)$ over $\lambda \geq 0$. We call this estimator $\hat{\lambda}_{CV}$. The solution to Problem (C) based on cross validation is computed by solving Problem (C) with $\lambda$ replaced by $\hat{\lambda}_{CV}$ and the $Y_i$'s replaced by the $\overline{Y}_i$'s; see Wahba [71, equations (2.4.23) and (2.4.24) on p. 33] for the details on how Problem (C) is solved using systems of linear equations.

We also solve Problem (C) with the $Y_i$'s replaced by the $\overline{Y}_i$'s and $\lambda$ replaced by $\lambda_n$ for $n \geq 1$. The specific choice of $(\lambda_n : n \geq 1)$ is given in Sections 6.5 and 6.6.

## 6.4. How to Select *m*

One needs a priori knowledge on $m$ to determine the right value of $m$. It should be noted that the smoothness of a function $g : \mathbb{R} \to \mathbb{R}$ is measured by $\int_{-\infty}^{\infty} \{g^{(2)}(x)\}^2 dx$. Thus, if $f_*$ itself is believed to be smooth, then we set $m = 2$. If $f_*$ is believed to have smooth partial derivatives of order 2, then we set $m = 4$.

## 6.5. *M/M*/1 Queue

We consider the case where $f_*(x)$ is the steady-state mean waiting time of a customer in a single-server queue under the first come, first served discipline with infinite capacity buffer, exponential service times, exponential interarrival times, unit arrival rate, and a service rate of $x \in [1.5, 2.0]$. $f_*(x)$ is known to be $1/x(x - 1)$; see Hillier and Lieberman [31, p. 781]. We chose an $M/M/1$ queue because there exists a closed-form formula for the steady-state waiting time of a customer, so we can compare our estimators with the true values. We set $X_i = 1.5 + i/(2n) - 1/(4n)$ for $1 \leq i \leq n$. For each $i \in \{1, \cdots, n\}$ and $j \in \{1, \cdots, r\}$ with $r = 100$, we generated $Y_{ij}$ by averaging the waiting times of the first 1,000 customers arriving at the queue when the queue is initialized empty and idle. We next computed the solution $\hat{g}_n$ to Problem (B) and the solution $\hat{f}_n$ to Problem (A) with the $Y_i$'s replaced by the $\overline{Y}_i$'s and $m = 4$. $S_n$ is estimated from (14), and $U_n$ is estimated from $J(\hat{g}_n)$. Both Problems (A) and (B) are solved using CVX.

For Problem (C), we computed $\hat{\lambda}_{CV}$ by minimizing $V(\lambda)$ in (15) over $\lambda \in \{10^{-11}, 5 \cdot 10^{-11}, 10^{-10}, 5 \cdot 10^{-10}, \cdots, 1\}$. In addition to $\hat{\lambda}_{CV}$, we used three sequences of real numbers, $\lambda(1) = (10^{-6}/n : n \geq 1)$, $\lambda(2) = (10^{-7}/n : n \geq 1)$, and $\lambda(3) = (10^{-8}/n : n \geq 1)$, for $\lambda$. To select $\lambda(2)$, we tried several sequences of real numbers and selected the one generating the lowest empirical integrated mean square error (EIMSE) values for $f_*$ and $f_*^{(2)}$. The rate of decay for $(\lambda_n : n \geq 1)$ was chosen to be of order $1/n$ based on the suggestion made by Cox [13].

To measure the accuracy of $\hat{g}_n$, the solution to Problem (B), we computed the EIMSE of $\hat{g}_n$ as follows:

$$\frac{1}{n} \sum_{i=1}^{n} (\hat{g}_n(X_i) - f_*(X_i))^2.$$

**Table 1.** The 95% confidence intervals of the EIMSE values for the estimators of $f_*$ computed from Problems (A)–(C) when $f_*$ is the steady-state mean waiting time of a customer in an $M/M/1$ queue. All values are measured in $10^{-4}$.

| $n$ | (A) | (B) | (C) | | | |
| --- | --- | --- | --- | --- | --- | --- |
| | | | CV | $\lambda(1)$ | $\lambda(2)$ | $\lambda(3)$ |
| 15 | $1.21 \pm 0.09$ | $1.14 \pm 0.09$ | $1.24 \pm 0.10$ | $1.13 \pm 0.09$ | $1.13 \pm 0.09$ | $1.16 \pm 0.09$ |
| 25 | $0.96 \pm 0.07$ | $0.87 \pm 0.06$ | $0.93 \pm 0.06$ | $0.86 \pm 0.06$ | $0.86 \pm 0.06$ | $0.86 \pm 0.06$ |
| 35 | $0.75 \pm 0.05$ | $0.67 \pm 0.05$ | $0.70 \pm 0.05$ | $0.65 \pm 0.04$ | $0.65 \pm 0.05$ | $0.67 \pm 0.05$ |

We also computed the EIMSE of $\hat{g}_n^{(2)}$ as follows:

$$\frac{1}{n}\sum_{i=1}^{n}(\hat{g}_n^{(2)}(X_i) - f_*^{(2)}(X_i))^2,$$

where $\hat{g}_n^{(2)}$ and $f_*^{(2)}$ denote the second derivatives of $\hat{g}_n$ and $f_*$, respectively. The EIMSE values of the solutions to Problems (A) and (C) and those of their second derivatives are computed similarly.

Table 1 reports the 95% confidence intervals of the EIMSE values of the estimators of $f_*$ computed from Problems (A)–(C) based on 400 iid replications for each $n$ value. The columns labeled as (A) and (B) report the results obtained from Problems (A) and (B), respectively. The columns labeled as CV, $\lambda(1)$, $\lambda(2)$, and $\lambda(3)$ report the results obtained from Problem (C) when $\lambda$ is set as $\hat{\lambda}_{CV}$, $\lambda(1)$, $\lambda(2)$, and $\lambda(3)$, respectively. The values in Table 1 are measured in $10^{-4}$. Table 2 reports the 95% confidence intervals of the EIMSE values of the estimators of $f_*^{(2)}$ computed from Problems (A)–(C) based on 400 iid replications for each $n$ value.

### 6.6. Stock Trader's Problem

We consider the stock trader's problem that motivated this paper. Consider a trader who trades stock options on a daily basis. Each day, his decision is either buy or sell a stock option, and he bases his decision on gamma, which is the second derivative of the value of the stock option as a function of the underlying stock price. The value of the stock option, $f_*$, typically has no closed-form formula, so simulation is required to estimate the value of the stock option and its second derivative with respect to the underlying stock price. Because simulation takes a long time to conduct, the estimation of $f_*$ and $f_*^{(2)}$ is done a day before any decision is made. Because $f_*$ and $f_*^{(2)}$ depend on the underlying stock price and because we cannot precisely predict the underlying stock price for the next day, $f_*$ and $f_*^{(2)}$ are typically estimated over a range of possible values of the underlying stock price. Hence, the trader's goal is to estimate gamma, $f_*^{(2)}$, over a range of possible values of the underlying stock price. Furthermore, in his estimation, he does not want to use any parameter that is not estimated from the data set.

To place this problem in a more specific setting, we consider the case where $f_*(x)$ is the price of a European call option on a nondividend-paying stock when the underlying stock price is $x \in [0, 2]$. We chose a European call option because there exists a closed-form formula for its price, so we can compare our estimates with the true values. Specifically, we consider the case where the strike price is $K = 1.3$, the risk-free annual interest rate is $s = 0.03$, the stock price volatility is 0.3, the underlying stock has a drift of 0.03, and the time to maturity of the option is $T = 1$ year. It is known that the price of this option is given by

$$f_*(x) = xN\left(\frac{\ln(x/K) + (s + \sigma^2/2)T}{\sigma\sqrt{T}}\right) - Ke^{-sT}N\left(\frac{\ln(x/K) + (s - \sigma^2/2)T}{\sigma\sqrt{T}}\right)$$

**Table 2.** The 95% confidence intervals of the EIMSE values for the estimators of $f_*^{(2)}$ computed from Problems (A)–(C) when $f_*$ is the steady-state mean waiting time of a customer in an $M/M/1$ queue.

| $n$ | (A) | (B) | (C) | | | |
| --- | --- | --- | --- | --- | --- | --- |
| | | | CV | $\lambda(1)$ | $\lambda(2)$ | $\lambda(3)$ |
| 15 | $7.52 \pm 0.42$ | $6.06 \pm 0.43$ | $19.63 \pm 3.70$ | $5.34 \pm 0.42$ | $5.32 \pm 0.44$ | $7.07 \pm 0.74$ |
| 25 | $6.79 \pm 0.36$ | $5.07 \pm 0.33$ | $14.73 \pm 2.85$ | $4.35 \pm 0.29$ | $4.21 \pm 0.33$ | $5.84 \pm 0.67$ |
| 35 | $5.87 \pm 0.29$ | $4.30 \pm 0.24$ | $9.38 \pm 1.77$ | $3.53 \pm 0.20$ | $3.28 \pm 0.24$ | $4.45 \pm 0.50$ |

**Table 3.** The 95% confidence intervals of the EIMSE values for the estimators of $f_*$ computed from Problems (A)–(C) when $f_*$ is the price of a European call option. All values are measured in $10^{-5}$.

| | | | (C) | | | |
|---|---|---|---|---|---|---|
| $n$ | (A) | (B) | CV | $\lambda(4)$ | $\lambda(5)$ | $\lambda(6)$ |
| 15 | $5.24 \pm 0.15$ | $1.72 \pm 0.11$ | $1.22 \pm 0.10$ | $1.23 \pm 0.08$ | $0.94 \pm 0.08$ | $1.02 \pm 0.08$ |
| 25 | $1.14 \pm 0.07$ | $1.14 \pm 0.07$ | $0.80 \pm 0.06$ | $0.82 \pm 0.04$ | $0.60 \pm 0.05$ | $0.66 \pm 0.05$ |
| 35 | $0.88 \pm 0.05$ | $0.88 \pm 0.05$ | $0.54 \pm 0.04$ | $0.60 \pm 0.03$ | $0.41 \pm 0.03$ | $0.46 \pm 0.03$ |

for $x \in [0, 2]$, where $N(\cdot)$ is the cumulative probability distribution function of a standard normal random variable; see, for example, Hull [32, p. 295].

To compute our estimators, we set $X_i = 2i/n - 1/n$ for $1 \le i \le n$. For each $X_i$, we assumed that the current stock price is $X_i$ and simulated a sample path $(S_t : 0 \le t \le T)$ of a geometric Brownian motion up to time $T$ with a drift of 0.03 and a volatility of 0.3. We then computed $Y_{ij}$, the price of the option by computing $\exp(-sT)\max(0, S_T - K)$, where $S_T$ is the value of the geometric Brownian motion we generated at time $T$. This procedure was repeated $r = 5{,}000$ times, generating $((X_i, Y_{ij}) : 1 \le i \le n, 1 \le j \le r)$. We next computed the solutions to Problems (A) and (B) with the $Y_i$'s replaced by the $\overline{Y}_i$'s and $m = 4$. $S_n$ is estimated from (14), and $U_n$ is estimated from $J(\hat{g}_n)$. Both Problems (A) and (B) are solved with CVX.

For Problem (C), we computed $\hat{\lambda}_{CV}$ by minimizing $V(\lambda)$ in (15) over $\lambda \in \{10^{-11}, 5 \cdot 10^{-11}, 10^{-10}, 5 \cdot 10^{-10}, \cdots, 1\}$. In addition to $\hat{\lambda}_{CV}$, we used three sequences of real numbers, $\lambda(4) = (10^{-5}/n : n \ge 1)$, $\lambda(5) = (10^{-6}/n : n \ge 1)$, and $\lambda(6) = (10^{-7}/n : n \ge 1)$, for $\lambda$. To select $\lambda(5)$, we tried several sequences of real numbers and selected the one generating the lowest EIMSE values for $f_*$ and $f_*^{(2)}$. The rate of decay for $(\lambda_n : n \ge 1)$ was chosen to be of order $1/n$ based on the suggestion made by Cox [13].

Table 3 reports the 95% confidence intervals of the EIMSE values of the estimators of $f_*$ computed from Problems (A)–(C) based on 400 iid replications for each $n$ value. The columns labeled as (A) and (B) report the results obtained from Problems (A) and (B), respectively. The columns labeled as CV, $\lambda(4)$, $\lambda(5)$, and $\lambda(6)$ report the results obtained from Problem (C) when $\lambda$ is set as $\hat{\lambda}_{CV}$, $\lambda(4)$, $\lambda(5)$, and $\lambda(6)$, respectively. The values in Table 3 are measured in $10^{-5}$.

Table 4 reports the 95% confidence intervals of the EIMSE values of the estimators of $f_*^{(2)}$ computed from Problems (A)–(C) based on 400 iid replications for each $n$ value. The values in Table 3 are measured in $10^{-1}$.

## 6.7. Observations from Numerical Experiments

When we compare Problems (A) and (B) with Problem (C) with the smoothing parameter estimated from cross validation, Problem (B) shows superior performance when estimating the second derivative of $f_*$. Problem (C) generates extreme estimates in a few percents of time, thereby degrading the overall performance. This phenomenon has been observed by other researchers; see the discussion in Wahba [71, p. 65].

When we compare Problems (A) and (B) with Problem (C) with a fixed smoothing parameter $\lambda_n$ for $n \ge 1$, Problem (C) shows superior performance. However, it should be noted that to select $\lambda_n$, we tried several sequences of real numbers and selected the one generating the lowest EIMSE values for $f_*$ and $f_*^{(2)}$. This was possible because we already knew what $f_*$ and $f_*^{(2)}$ were in all of our numerical examples. In reality, one does not have a priori knowledge of the exact values of $f_*(X_i)$ and $f_*^{(2)}(X_i)$ for $1 \le i \le n$, so finding $(\lambda_n : n \ge 1)$ this way is not feasible. However, in our numerical experiments, the results from Problem (C) with a fixed smoothing parameter $\lambda_n$ give us a sense of how well Problems (A) and (B) perform compared with the best possible performance of Problem (C). Tables 1–4 report that the performance of Problems (A) and (B) is as good as that of Problem (C) with the best possible choice of $(\lambda_n : n \geq 1)$.

**Table 4.** The 95% confidence intervals of the EIMSE values for the estimators of $f_*^{(2)}$ computed from Problems (A)–(C) when $f_*$ is the price of a European call option. All values are measured in $10^{-1}$.

| | | | (C) | | | |
|---|---|---|---|---|---|---|
| $n$ | (A) | (B) | CV | $\lambda(4)$ | $\lambda(5)$ | $\lambda(6)$ |
| 15 | $0.96 \pm 0.03$ | $0.57 \pm 0.05$ | $1.72 \pm 0.32$ | $0.46 \pm 0.01$ | $0.36 \pm 0.04$ | $0.90 \pm 0.12$ |
| 25 | $0.44 \pm 0.03$ | $0.45 \pm 0.03$ | $2.80 \pm 0.58$ | $0.35 \pm 0.01$ | $0.25 \pm 0.03$ | $0.68 \pm 0.09$ |
| 35 | $0.36 \pm 0.01$ | $0.36 \pm 0.01$ | $2.89 \pm 0.66$ | $0.28 \pm 0.01$ | $0.18 \pm 0.02$ | $0.53 \pm 0.06$ |

**Table 5.** The 95% confidence intervals of the EIMSE values for $\hat{g}_n$ and $\hat{g}_n^{(2)}$ when $S_n$ was set as the average of sample variances across five intervals.

| $n$ | EIMSE for $\hat{g}_n$ | EIMSE for $\hat{g}_n^{(2)}$ |
|---|---|---|
| 30 | $0.0028 \pm 0.0001$ | $6.68 \pm 0.42$ |
| 40 | $0.0021 \pm 0.0001$ | $5.36 \pm 0.37$ |
| 50 | $0.0016 \pm 0.0001$ | $4.03 \pm 0.28$ |

## 6.8. When Multiple Observations Cannot Be Made at a Fixed Design Point in $[a, b]^d$

Throughout Section 6, we assumed that multiple observations can be made at any fixed design point $\mathbf{x}$ in $[a, b]^d$. These multiple observations were used to empirically estimate $S_n$. In many applications, multiple observations at a point $\mathbf{x} \in [a, b]^d$ may not be available. In such cases, one needs to use other ways to empirically estimate $S_n$. One possible approach is partitioning $[a, b]^d$ into a finite number of hyperrectangles, computing the sample variance of the $Y_i$'s whose $X_i$'s fall within each hyperrectangle, and computing the average of these sample variances (weighted by the density function of $X$ evaluated at each hyperrectangle) as an estimate of $S_n$. This approach is based on the observation that $S_n$ should ideally be $\mathbb{E}(\text{Var}(Y|X))$.

To illustrate this approach numerically, we consider the following example: $f_* : [0, 1] \to \mathbb{R}$ is given by $f_*(x) = (x - 1/4)^2$ for $x \in [0, 1]$, $m = 4$, $X_i = i/n - 1/(2n)$ for $1 \leq i \leq n$, and $Y_i = f_*(X_i) + \varepsilon_i$ for $1 \leq i \leq n$, where the $\varepsilon_i$'s follow a uniform distribution over $[-0.25, 0.25]$. We assumed that only one observation $Y_i = f_*(X_i) + \varepsilon_i$ is made for $1 \leq i \leq n$. To estimate $S_n$, we partitioned $[0, 1]$ into the five intervals $I^1 = [0, 0.2], I^2 = [0.2, 0.4], I^3 = [0.4, 0.6], I^4 = [0.6, 0.8]$, and $I^5 = [0.8, 1.0]$ and computed the sample variances, say $S^1, S^2, S^3, S^4$, and $S^5$, in which $S^k$ is the sample variance of the $Y_i$'s whose $X_i$'s fall in $I^k$ for $1 \leq k \leq 5$. We set $S_n$ equal to the average of $S^1, S^2, S^3, S^4$, and $S^5$. We then solved Problem (B) to obtain $\hat{g}_n$ and $\hat{g}_n^{(2)}$. This procedure was repeated 1,600 times independently. Using these 1,600 replications, we computed the 95% confidence intervals of the EIMSE values of $\hat{g}_n$ and $\hat{g}_n^{(2)}$, respectively. Table 5 reports the 95% confidence intervals for a variety of $n$ values. The EIMSE values in Table 5 decrease as $n$ increases, indicating that the choice of $S_n$ is appropriate.

# 7. Conclusions

We conclude this paper with some discussions on future research topics.

## 7.1. How to Choose $S_n$ When Multiple Observations Are Not Available

In Section 6.8, we discussed a way to empirically estimate $S_n$ when multiple observations are not available at a design point in $[a, b]^d$. In this approach, we divide $[a, b]^d$ into smaller hyperrectangles, compute the sample variance of the $Y_i$'s whose $X_i$'s fall within each hyperrectangle, and compute the average of these sample variances (weighted by the density function of $X$ evaluated at each hyperrectangle) as an estimate of $S_n$. We conjecture that this approach will generate an asymptotically consistent estimator if the volume of the hyperrectangles decreases to zero and the number of $X_i$'s within each hyperrectangle increases to infinity as $n \to \infty$. A more rigorous study on this conjecture is a good future research topic.

## 7.2. Combining the Smoothness Condition with Shape Constraints

Because our proposed formulations, Problems (A) and (B), impose only the smoothness condition on the underlying function $f_*$, the proposed estimators do not necessarily produce a function that satisfies additional shape conditions on $f_*$, such as convexity or monotonicity. For example, even if $f_*$ is convex, our estimators are not necessarily convex. Figure 1 shows the solution to Problem (B) when $f_* : [0, 1] \to \mathbb{R}$ is given by $f_*(x) = (x - 1/4)^2$ for $x \in [0, 1]$, $m = 4$, $n = 10$, $X_i = i/n - 1/(2n)$ for $1 \leq i \leq n$, and $Y_i = f_*(X_i) + \varepsilon_i$ for $1 \leq i \leq n$, where the $\varepsilon_i$'s follow a uniform distribution over $[-0.5, 0.5]$ and $S_n = 1/12$. Even if $f_*$ is convex, $\hat{g}_n$ is not necessarily convex.

One possible fix to this phenomenon is incorporating additional conditions to our formulation. A good topic for future research is combining shape constraints, such as convexity and monotonicity, with the smoothness condition in our proposed formulations.

**Figure 1.** The solid line is $f_*$, the dots are the $Y_i$'s, and the dashed line is the solution to Problem (B).

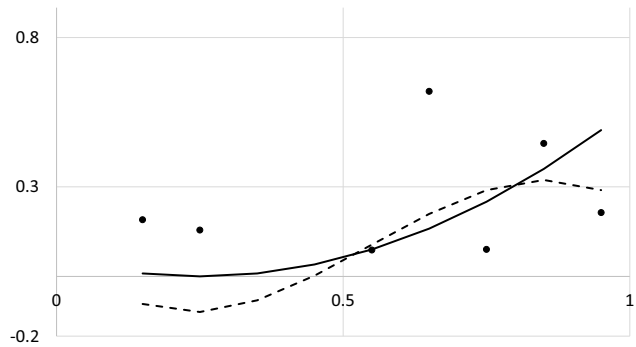

## 7.3. Establishing Consistency Under a Condition That Is Easier to Verify

Corollary 2 is established under Assumption 7, which is hard to verify. We conjecture that the conclusions of Corollary 2 hold under a more natural condition that $S_n \to \sigma^2$ as $n \to \infty$. This remains as an open question. Another open question is computing the convergence rate of $\hat{g}_n$ under suitable conditions on $S_n$.

## 7.4. Extending Our Formulations to Shape-Constrained Estimation Problems

Problem (A) can be seen as an estimation problem of a smooth function by minimizing the sum of squared errors under the smoothness condition $J(f) \le U_n$. This type of formulation has been used for shape-constrained estimation problems, such as convex regression. For example, Mazumder et al. [49] proposed estimating an unknown convex function by minimizing the sum of squared errors over convex functions satisfying a Lipschitz condition. An interesting question for future research is how to extend Problem (B) to shape-constrained problems, such as convex and isotonic regression problems.

## Appendix A. Proofs of Propositions 1–7 and Lemmas 1 and 2

To establish the existence of the solutions to Problems (A) and (B), we define the following functions $\varphi_n : \mathbb{R}^n \to \mathbb{R}$ and $\phi_n : \mathbb{R}^n \to \mathbb{R}$. For any $\mathbf{z} = (z_1, \cdots, z_n) \in \mathbb{R}^n$, consider the following problem:

Problem (I) : Minimize$_{f \in \mathcal{F}_m} J(f)$ subject to $f(X_i) = z_i$ $\quad 1 \le i \le n$.

Under Assumption 1, the solution to Problem (I) exists uniquely; see Meinguet [50, equation 29 and theorem 3 on p. 299]. We will denote the solution to Problem (I) by $f_\mathbf{z}$. Define $\varphi_n(\mathbf{z}) = J(f_\mathbf{z})$. On the other hand, $\phi_n : \mathbb{R}^n \to \mathbb{R}$ is defined by $\phi_n(\mathbf{z}) = (1/n) \sum_{i=1}^{n} (Y_i - z_i)^2$ for $\mathbf{z} = (z_1, \cdots, z_n) \in \mathbb{R}^n$. It can be easily seen that $\varphi_n$ is convex over $\mathbb{R}^n$ and that $\phi_n$ is strictly convex over $\mathbb{R}^n$.

For any $S \ge 0$ and $U \ge 0$, we define $\mathcal{V}_U \subset \mathbb{R}^n$ and $\mathcal{T}_S \subset \mathbb{R}^n$ by

$$\mathcal{V}_U = \{\mathbf{z} \in \mathbb{R}^n : \varphi_n(\mathbf{z}) \le U\} \quad \text{and} \quad \mathcal{T}_S = \{\mathbf{z} \in \mathbb{R}^n : \phi_n(\mathbf{z}) \le S\}.$$

Both $\mathcal{S}_U$ and $\mathcal{T}_S$ are nonempty, closed, and convex subsets of $\mathbb{R}^n$.

We recall that $\Psi_n : [0, \infty) \to \mathbb{R}$ is defined as follows. For any nonnegative real number $u$, $\Psi_n(u) = E_n(f^u)$, where $f^u$ is a solution Problem (A) with $U_n = u$.

**Proof of Proposition 1.** Because $\mathcal{V}_U$ is nonempty, closed, and convex and because $\phi_n$ is strictly convex over $\mathcal{V}_U$, there exists a unique minimizer $\mathbf{z}^* = (z_1^*, \cdots, z_n^*)$ of $\phi_n$ over $\mathcal{V}_U$ and $f_{\mathbf{z}^*}$ is a solution to Problem (A).

To prove the second part, let $\hat{f}_n$ and $\tilde{f}_n$ be two solutions to Problem (A). Then, $(\hat{f}_n(X_1), \cdots, \hat{f}_n(X_n))$ and $(\tilde{f}_n(X_1), \cdots, \tilde{f}_n(X_n))$ are two minimizers of $\phi_n$ over $\mathcal{V}_U$. By the uniqueness of the minimizer of $\phi_n$ over $\mathcal{V}_U$, it follows that $\hat{f}_n(X_i) = \tilde{f}_n(X_i)$ for $1 \le i \le n$. $\quad \square$

**Proof of Proposition 2.** Because $\mathcal{T}_S$ is nonempty, closed, and convex and because $\varphi_n$ is convex, there exists a minimizer $\mathbf{z}^* = (z_1^*, \cdots, z_n^*)$ of $\varphi_n$ over $\mathcal{T}_S$ and $f_{\mathbf{z}^*}$ is a solution to Problem (B).

To prove the second part, we note that for any $f \in \mathcal{F}_m$, $J(f) = \|f\|_m^2$, and $\|\cdot\|_m$ is a seminorm in $\mathcal{F}_m$ and a norm in $\mathcal{F}_m/\mathcal{P}_{m-1}$. We will denote an equivalence class in $\mathcal{F}_m/\mathcal{P}_{m-1}$ by $[f]$ for $f \in \mathcal{F}_m$.

Let $\mathbf{y}^*$ and $\mathbf{z}^*$ be minimizers of $\varphi_n$ over $\mathcal{T}_S$. Because

$$\||[f_{1/2\mathbf{y}^*+1/2\mathbf{z}^*}]\||_m \le \||1/2[f_{\mathbf{y}^*}] + 1/2[f_{\mathbf{z}^*}]\||_m \le (1/2)\||[f_{\mathbf{y}^*}]\||_m + (1/2)\||[f_{\mathbf{z}^*}]\||_m = \||[f_{\mathbf{y}^*}]\||_m$$

and $\||[f_{\mathbf{y}^*}]\||_m \le \||[f_{1/2\mathbf{y}^*+1/2\mathbf{z}^*}]\||_m$, we can conclude

$$\||1/2[f_{\mathbf{y}^*}] + 1/2[f_{\mathbf{z}^*}]\||_m = (1/2)\||[f_{\mathbf{y}^*}]\||_m + (1/2)\||[f_{\mathbf{z}^*}]\||_m,$$

which implies either $[f_{\mathbf{y}^*}] = 0$ or $[f_{\mathbf{y}^*}] = c[f_{\mathbf{z}^*}]$ for some constant $c$. In the latter case, $c$ must be equal to one because we have $\||f_{\mathbf{y}^*}\||_m = \||f_{\mathbf{z}^*}\||_m$. In either case, $f_{\mathbf{y}^*} - f_{\mathbf{z}^*} \in \mathcal{P}_{m-1}$. $\quad \square$

**Proof of Lemma 1.** The existence of $f_P$ follows from the fact that

$$\frac{1}{n}\sum_{i=1}^{n}(Y_i - (c_1 p_1(X_i) + \cdots + c_M p_M(X_i)))^2 \tag{A.1}$$

is convex over $(c_1, \cdots, c_M) \in \mathbb{R}^M$.

The only nontrivial part is that $\Psi_n$ is strictly decreasing over $[0, J(f_I)]$. Let $0 \le x < y \le J(f_I)$ be given. We need to show that $\Psi_n(x) > \Psi_n(y)$. Suppose, on the contrary, that

$$\Psi_n(x) = \Psi_n(y). \tag{A.2}$$

Let $f^x$ and $f^y$ be the solutions to Problem (A) with $U = x$ and $U = y$, respectively. Note that $(f^x(X_1), \cdots, f^x(X_n)) \in \mathcal{V}_y$ and minimizes $\phi_n$ over $\mathcal{V}_y$ because of (A.2).

We will construct an $n$-dimensional ball around $(f^x(X_1), \cdots, f^x(X_n))$ that is contained in $\mathcal{V}_y$. If this is proven, it will follow that $f^x(X_i) = Y_i$ for $1 \le i \le n$, and hence, $f^x$ interpolates the $Y_i$'s, so $J(f_I) \le J(f^x)$. Together with the facts $x < J(f_I)$ and $J(f^x) \le x$, we will reach a contradiction.

To construct such a ball, we notice that the $X_i$'s are distinct, so we can find $\delta_0 > 0$ such that $\|X_i - X_j\| > \delta_0$ for $i \ne j$. For any $\epsilon \in (0, \epsilon_0]$, we will show that $(f^x(X_1) + \epsilon, \cdots, f^x(X_n) + \epsilon)$ is in $\mathcal{V}_y$. ($\epsilon_0 > 0$ will be specified later.) Define $\theta_\epsilon : \mathbb{R}^n \to \mathbb{R}$ by

$$\theta_\epsilon(\mathbf{z}) = f^x(\mathbf{z}) + \epsilon(\chi_1(\mathbf{z}) + \cdots + \chi_n(\mathbf{z}))$$

for $\mathbf{z} \in \mathbb{R}^n$, where the $\chi_i$'s are the "mollifiers" defined by

$$\chi_i(\mathbf{z}) = \begin{cases} e \cdot \exp(\delta_0/(\|\mathbf{z} - X_i\|^2 - \delta_0)), & \text{if } \|\mathbf{z} - X_i\| < \delta_0 \\ 0, & \text{otherwise} \end{cases}$$

for $\mathbf{z} \in \mathbb{R}^n$ and $1 \le i \le n$. The $\chi_i$'s have the following properties:
  i. $\chi_i(X_i) = 1$ and $\chi_i(X_j) = 0$ for all $j \ne i$;
  ii. the $\chi_i$'s are infinitely differentiable; and
  iii. $J(\chi_i) \le C$ for all $1 \le i \le n$ and some constant $C$.
It follows that $\theta_\epsilon(X_i) = f^x(X_i) + \epsilon$ for $1 \le i \le n$, and

$$J(\theta_\epsilon) = \|\theta_\epsilon\|_m^2 \le (\|f^x\|_m + \epsilon\|\chi_1 + \cdots + \chi_n\|_m)^2$$
$$\le J(f^x) + 2\epsilon n C^{1/2}\|f^x\|_m + \epsilon^2 n^2 C,$$

so taking $\epsilon_0$ sufficiently small yields $J(\theta_\epsilon) \le J(f^y) \le y$. Thus, $(f^x(X_1) + \epsilon, \cdots, f^x(X_n) + \epsilon)$ is in $\mathcal{V}_y$.  $\square$

**Proof of Proposition 4.** Let $0 < S_n \le E_n(f_P)$ be given. By Proposition 2, there exists a solution $\hat{g}_n$ to Problem (B). Let $U^* = \Psi_n^{-1}(S_n)$ and $\tilde{h}$ be a solution to Problem (A) with $U_n = U^*$. We will first show that $\hat{g}_n$ is also a solution to Problem (A) with $U_n = U^*$ and satisfies $E_n(\hat{g}_n) = S_n$. By definition, $\tilde{h}$ minimizes $E_n(f)$ over all functions $f$ satisfying $J(f) \le U^*$ and satisfies

$$E_n(\tilde{h}) = S_n \tag{A.3}$$

and

$$J(\tilde{h}) \le U^*. \tag{A.4}$$

By (A.3), $\tilde{h}$ is a feasible solution to Problem (B), so we should have

$$J(\hat{g}_n) \le J(\tilde{h}). \tag{A.5}$$

By (A.4) and (A.5), we have

$$J(\hat{g}_n) \le U^*, \tag{A.6}$$

and hence, $\hat{g}_n$ is a feasible solution to Problem (A) with $U_n = U^*$. Thus,

$$E_n(\tilde{h}) \le E_n(\hat{g}_n). \tag{A.7}$$

Also, $\hat{g}_n$ is a feasible solution to Problem (B), so we should have

$$E_n(\hat{g}_n) \le S_n. \tag{A.8}$$

By (A.3), (A.7), and (A.8), we can conclude

$$E_n(\hat{g}_n) = E_n(\tilde{h}) = S_n,$$

and hence, together with (A.6), $\hat{g}_n$ becomes a solution to Problem (A) with $U_n = U^*$.

Next, we turn to the uniqueness of the solution to Problem (B). Let $\hat{g}_n$ and $\tilde{g}_n$ be two solutions to Problem (B). By Proposition 1, we have

$$\hat{g}_n - \tilde{g}_n = \tilde{p}_n \tag{A.9}$$

for some $\tilde{p}_n \in \mathcal{P}_{m-1}$. By the previous arguments, $\hat{g}_n$ and $\tilde{g}_n$ are also solutions to Problem (A) with $U_n = U^*$. By Proposition 1, we have

$$\hat{g}_n(X_i) = \tilde{g}_n(X_i) \tag{A.10}$$

for $1 \le i \le n$. From (A.9) and (A.10), we have $\tilde{p}_n(X_i) = 0$ for $1 \le i \le n$, which implies $\tilde{p}_n(\mathbf{x}) = 0$ for all $\mathbf{x} \in \mathbb{R}^d$ by the $\mathcal{P}_{m-1}$-unisolvency of the $X_i$'s. Hence, $\hat{g}_n(\mathbf{x}) = \tilde{g}_n(\mathbf{x})$ for all $\mathbf{x} \in \mathbb{R}^d$. $\square$

**Proof of Proposition 3.** Let $0 \le U_n < J(f_l)$ be given. By Proposition 1, there exists a solution $\hat{f}_n$ to Problem (A). Let $S^* = \Psi_n(U_n)$ and $\tilde{h}$ be a solution to Problem (B) with $S_n = S^*$. We will first show that $\hat{f}_n$ is also a solution to Problem (B) with $S_n = S^*$ and satisfies $J(\hat{f}_n) = U_n$. First, we will prove that

$$U_n \le J(\tilde{h}). \tag{A.11}$$

To prove (A.11), we note that by Proposition 4, $\tilde{h}$ is a solution to

$$\text{minimize}_{f \in \mathcal{F}_m} E_n(f) \quad \text{subject to } J(f) \le U_n$$

with $E_n(\tilde{h}) = S^*$. If, on the contrary, $J(\tilde{h}) < U_n$, then $\tilde{h}$ is also a solution to

$$\text{minimize}_{f \in \mathcal{F}_m} E_n(f) \quad \text{subject to } J(f) \le J(\tilde{h})$$

with $E_n(\tilde{h}) = S^*$, which contradicts the fact that $\Psi_n$ is strictly decreasing over $[0, J(f_l)]$. Thus, (A.11) is proven.

By Proposition 4, $\tilde{h}$ is a solution to Problem (A), so we have

$$E_n(\hat{f}_n) = E_n(\tilde{h}) = S^*. \tag{A.12}$$

By (A.12), $\hat{f}_n$ is a feasible solution to Problem (B) with $S_n = S^*$, so we should have

$$J(\tilde{h}) \le J(\hat{f}_n). \tag{A.13}$$

Because $\hat{f}_n$ is a feasible solution to Problem (A), we have

$$J(\hat{f}_n) \le U_n. \tag{A.14}$$

By combining (A.11), (A.13), and (A.14), we have

$$J(\tilde{h}) = J(\hat{f}_n) = U_n. \tag{A.15}$$

By (A.12) and (A.15), $\hat{f}_n$ is also a solution to Problem (B) with $S = S^*$ and satisfies $J(\hat{f}_n) = U_n$.

We next turn to the uniqueness of the solution to Problem (A). Let $\hat{f}_n$ and $\tilde{f}_n$ be two solutions to Problem (A). By the previous arguments, $\hat{f}_n$ and $\tilde{f}_n$ are also two solutions to Problem (B) with $S_n = S^*$. By Proposition 4, the solution to Problem (B) with $S_n = S^*$ is unique, so $\hat{f}_n = \tilde{f}_n$. $\square$

**Proof of Proposition 5.** Combine Propositions 3 and 4. $\square$

**Proof of Proposition 6.** Let $\mathcal{H}_0 = \text{span}\{p_1, \cdots, p_M\}$ and $\mathcal{H}_1$ be the reproducing kernel Hilbert space with the $R(\mathbf{s}, \mathbf{t})$'s as the reproducing kernels. Then, $\mathcal{F}_m$ is the direct sum of $\mathcal{H}_0$ and $\mathcal{H}_1$; see Meinguet [50, equation 13 on p. 295 and equation 20 on p. 296]. Any element $f$ of $\mathcal{F}_m$ has the following representation:

$$f(\mathbf{x}) = \sum_{i=1}^{M} c_i p_i(\mathbf{x}) + \sum_{i=1}^{n} d_i R(\mathbf{x}, X_i) + \rho(\mathbf{x})$$

for $\mathbf{x} \in \mathbb{R}^d$, where $\rho$ is some element in $\mathcal{F}_m$ perpendicular to $p_1(\cdot), \cdots, p_M(\cdot), R(\cdot, X_1), \cdots, R(\cdot, X_n)$ because of the property of Hilbert spaces. Let $h(\cdot) = \sum_{i=1}^{n} d_i R(\cdot, X_i) + \rho(\cdot)$. By Meinguet [50, equation 14 on p. 295], for $j \in \{1, 2, \cdots, n\}$,

$$h(X_j) = (R(X_j, \cdot), h(\cdot))_m$$

$$= (R(X_j, \cdot), \sum_{i=1}^{n} d_i R(\cdot, X_i) + \rho(\cdot))_m$$

$$= \sum_{i=1}^{n} d_i R(X_j, X_i)$$

because $\rho$ is perpendicular to $R(\cdot, X_1), \cdots, R(\cdot, X_n)$ and $R(X_j, \cdot)$ is the representer of evaluation at $X_j$. Hence, Problem (A) can be expressed as $\text{Minimize}_{f \in \mathcal{F}_m} \frac{1}{n} \sum_{i=1}^{n} (Y_i - (\sum_{j=1}^{M} c_j p_j(X_i) + \sum_{j=1}^{n} d_j R(X_j, X_i)))^2$ subject to $J(\sum_{i=1}^{n} d_i R(\cdot, X_i)) + J(\rho) \le U_n$. Hence, if $f(\mathbf{x}) = \sum_{i=1}^{M} c_i p_i(\mathbf{x}) + \sum_{i=1}^{n} d_i R(\mathbf{x}, X_i) + \rho(\mathbf{x})$ is a solution to Problem (A), then $g(\mathbf{x}) = \sum_{i=1}^{M} c_i p_i(\mathbf{x}) + \sum_{i=1}^{n} d_i R(\mathbf{x}, X_i)$ is also a solution to

Problem (A). Also, for such $g$, note that

$$
\begin{aligned}
J(g) &= J\left(\sum_{i=1}^n d_i R(\cdot, X_i)\right) \\
&= \left(\sum_{i=1}^n d_i R(\cdot, X_i), \sum_{i=1}^n d_i R(\cdot, X_i)\right)_m \\
&= \sum_{i=1}^n \sum_{j=1}^n d_i d_j (R(\cdot, X_i), R(\cdot, X_j))_m \\
&= \sum_{i=1}^n \sum_{j=1}^n d_i d_j R(X_i, X_j). \quad \square
\end{aligned}
$$

**Proof of Proposition 7.** Let $\mathcal{H}_0 = \text{span}\{p_1, \cdots, p_M\}$ and $\mathcal{H}_1$ be the reproducing kernel Hilbert space with the $R(\mathbf{s}, \mathbf{t})$'s as the reproducing kernels. Then, $\mathcal{F}_m$ is the direct sum of $\mathcal{H}_0$ and $\mathcal{H}_1$; see Meinguet [50, equation 13 on p. 295 and equation 20 on p. 296]. Any element $f$ of $\mathcal{F}_m$ has the following representation:

$$
f(\mathbf{x}) = \sum_{i=1}^M c_i p_i(\mathbf{x}) + \sum_{i=1}^n d_i R(\mathbf{x}, X_i) + \rho(\mathbf{x})
$$

for $\mathbf{x} \in \mathbb{R}^d$, where $\rho$ is some element in $\mathcal{F}$ perpendicular to $p_1(\cdot), \cdots, p_M(\cdot), R(\cdot, X_1), \cdots, R(\cdot, X_n)$ because of the property of Hilbert spaces. By arguments similar to those in the proof of Proposition 6, Problem (B) can be expressed as $\text{Minimize}_{f \in \mathcal{F}} J(\sum_{i=1}^n d_i R(\cdot, X_i)) + J(\rho)$ subject to $\frac{1}{n} \sum_{i=1}^n (Y_i - (\sum_{j=1}^M c_j p_j(X_i) + \sum_{j=1}^n d_j R(X_i, X_j)))^2 \le S_n$. Hence, $J(\rho) = 0$ and $\rho = 0$. The rest of the proposition follows from arguments similar to those in the proof of Proposition 6. $\quad \square$

**Proof of Lemma 2.** From the assumption that $\tau$ is continuous on $[a, b]^d$, the $X_i$'s are mutually distinct a.s. By Chui [10, theorem 9.4 on p. 135] or Chung and Yao [11], there exists a $\mathcal{P}_{m-1}$-unisolvent set $\{\mathbf{x}_1^*, \cdots, \mathbf{x}_r^*\}$ in $[a, b]^d$ with some fixed positive integer $r$.

Define a function $H : \mathbb{R}^{r \times d} \to \mathbb{R}$ by

$$
H(\mathbf{x}_1, \cdots, \mathbf{x}_r) = \det(P_{\mathbf{x}_1, \cdots, \mathbf{x}_r}^T P_{\mathbf{x}_1, \cdots, \mathbf{x}_r})
$$

for $(\mathbf{x}_1, \cdots, \mathbf{x}_r) \in \mathbb{R}^{r \times d}$, where

$$
P_{\mathbf{x}_1, \cdots, \mathbf{x}_r} = \begin{pmatrix} p_1(\mathbf{x}_1) & \cdots & p_M(\mathbf{x}_1) \\ \vdots & & \vdots \\ p_1(\mathbf{x}_r) & \cdots & p_M(\mathbf{x}_r) \end{pmatrix}.
$$

Then, $H(\mathbf{x}_1, \cdots, \mathbf{x}_r) \ne 0$ at $(\mathbf{x}_1, \cdots, \mathbf{x}_r) = (\mathbf{x}_1^*, \cdots, \mathbf{x}_r^*)$ because $P_{\mathbf{x}_1^*, \cdots, \mathbf{x}_r^*}^T P_{\mathbf{x}_1^*, \cdots, \mathbf{x}_r^*}$ is positive definite. Because $H$ is continuous, there exists $\epsilon_0 > 0$ such that

$$
\|(\mathbf{x}_1, \cdots, \mathbf{x}_r) - (\mathbf{x}_1^*, \cdots, \mathbf{x}_r^*)\| \le \epsilon_0
$$

implies $H(\mathbf{x}_1, \cdots, \mathbf{x}_r) \ne 0$. Note that $H(\mathbf{x}_1, \cdots, \mathbf{x}_r) \ne 0$ implies that $\{\mathbf{x}_1, \cdots, \mathbf{x}_r\}$ is a $\mathcal{P}_{m-1}$-unisolvent. By Assumption 2, there exists a subset $\{X_1^*, \cdots, X_r^*\} \subset \{X_1, \cdots, X_n\}$ such that

$$
\|X_i^* - \mathbf{x}_i^*\| < \epsilon_0/r \tag{A.16}
$$

for $1 \le i \le r$ for all $n$ sufficiently large a.s. We note that (A.16) implies

$$
\|(X_1^*, \cdots, X_r^*) - (\mathbf{x}_1^*, \cdots, \mathbf{x}_r^*)\| \le \epsilon_0,
$$

and hence, $\{X_1^*, \cdots, X_r^*\}$ is a $\mathcal{P}_{m-1}$-unisolvent set. $\quad \square$

## Appendix B. Proofs of Theorems 1–4 and Corollaries 1–3
We start by defining a set of functions that will play an important role in the proofs of Theorems 1–4. We define $\mathcal{A}_m$ by

$$
\mathcal{A}_m = \left\{ f \in \mathcal{F}_m : \int_{[a,b]^d} \{D^\alpha f(\mathbf{x})\}^2 d\mathbf{x} \le c_A + c_B + 1 \text{ for all } \alpha \text{ such that} \right.
$$

$$
\left. |\alpha| = m, |f(\mathbf{x})| \le \alpha_0, |f(\mathbf{x}) - f(\mathbf{y})| \le \alpha_1 \|\mathbf{x} - \mathbf{y}\|^\delta \text{ for } \mathbf{x}, \mathbf{y} \in [a, b]^d \right\} \tag{B.1}
$$

for some constants $\alpha_0$ and $\alpha_1$, where $\delta = 1$ when $m \geq 2$ and $\delta = 1/2$ when $m = 1$. We endow $\mathcal{A}_m$ with a metric $d_\infty$ and a pseudo-metric $d_n$ defined as follows:

$$d_\infty(f_1, f_2) = \sup_{\mathbf{x} \in [a,b]^d} |f_1(\mathbf{x}) - f_2(\mathbf{x})|$$

and

$$d_n(f_1, f_2) = \left\{ \frac{1}{n} \sum_{i=1}^{n} (f_1(X_i) - f_2(X_i))^2 \right\}^{1/2}$$

for $f_1, f_2 \in \mathcal{A}_m$.

In Lemmas B.1 and B.2, we will establish that Assumption 7 guarantees that $\hat{g}_n$, restricted to $[a,b]^d$, belongs to $\mathcal{A}_m$ for $n$ sufficiently large a.s. In the proof of Lemma B.3, we will use the fact that $\mathcal{A}_m$ can be covered by a finite number of balls with radius $\epsilon$ in metric $d_\infty$ for any $\epsilon > 0$. In the proof of Theorem 3, we will use the fact that the number of balls of radius $\epsilon$ needed to cover $\mathcal{A}_m$ in metric $d_n$ is of order $\exp(\epsilon^{-d/m})$.

To prove Theorems 1 and 2, we need Lemmas B.1, B.2, B.3, and B.4. Here is our first lemma.

**Lemma B.1.** *Let $\Omega$ be an open-bounded subset of $\mathbb{R}^d$ containing $[a,b]^d$. Under Assumptions 2, 3, 7, and 9, there exists a constant $\alpha_0$, depending on $\Omega$, such that*

$$\sup_{\mathbf{x} \in \Omega} |\hat{g}_n(\mathbf{x})| \leq \alpha_0$$

*for $n$ sufficiently large a.s.*

**Proof of Lemma B.1.** The proof is divided into three steps.

Step 1. We prove $(1/n) \sum_{i=1}^{n} \hat{g}_n(X_i)^2$ is bounded for $n$ sufficiently large (uniformly on $n$) a.s. To see this, one notices that by Assumption 3, Assumption 9, and the strong law of large numbers, we have

$$\frac{1}{n} \sum_{i=1}^{n} \hat{g}_n(X_i)^2 \leq \frac{2}{n} \sum_{i=1}^{n} (\hat{g}_n(X_i) - Y_i)^2 + \frac{2}{n} \sum_{i=1}^{n} Y_i^2$$

$$\leq \frac{2}{n} \sum_{i=1}^{n} (\hat{g}_n(X_i) - f_*(X_i))^2 - \frac{4}{n} \sum_{i=1}^{n} \varepsilon_i(\hat{g}_n(X_i) - f_*(X_i)) + \frac{2}{n} \sum_{i=1}^{n} \varepsilon_i^2 + \frac{2}{n} \sum_{i=1}^{n} Y_i^2$$

$$\leq 2B_n + \frac{2}{n} \sum_{i=1}^{n} \varepsilon_i^2 + \frac{2}{n} \sum_{i=1}^{n} Y_i^2 \leq 2\sigma^2 + 2\mathbb{E}(f_*(X) + \varepsilon)^2 + 1 \triangleq M_1 \tag{B.2}$$

for $n$ sufficiently large a.s. We used the fact that $f_* \in \mathcal{F}_m$ and $2m > d$, and hence, $f_*$ is continuous on $\mathbb{R}^d$ and is bounded over $[a,b]^d$.

Step 2. We use the Sobolev integral identity to express $\hat{g}_n$ as the sum of a polynomial of degree less than $m$ and a bounded function over $\Omega$. To fill in the details, note that $\hat{g}_n$ is continuous over $\mathbb{R}^d$ and hence, belongs to $L_2(\Omega)$ when it is restricted to $\Omega$. By Assumption 7 and Adams and Fournier [1, theorem 2 on p. 717], $\hat{g}_n$, restricted to $\Omega$, belongs to $W_m(\Omega)$ for $n$ sufficiently large a.s. The Sobolev integral identity (see Oden and Reddy [54, theorem 3.6 on p. 78]) allows us to express

$$\hat{g}_n(\mathbf{x}) = \hat{u}_n(\mathbf{x}) + \hat{v}_n(\mathbf{x})$$

for all $\mathbf{x} \in \Omega$, where $\hat{u}_n$ is a polynomial of degree less than $m$,

$$\hat{v}_n(\mathbf{x}) = \int_\Omega \|\mathbf{x} - \mathbf{y}\|^{m-d} \sum_{|\alpha|=m} Q_\alpha(\mathbf{x}, \mathbf{y}) D^\alpha \hat{g}_n(\mathbf{y}) d\mathbf{y}$$

for $\mathbf{x} \in \Omega$, and $Q_\alpha(\mathbf{x}, \mathbf{y})$, $|\alpha| = m$, are bounded infinitely differentiable functions of $\mathbf{x}$ and $\mathbf{y}$. Hölder's inequality implies

$$|\hat{v}_n(\mathbf{x})| \leq \sum_{|\alpha|=m} \left\{ \int_\Omega \|\mathbf{x} - \mathbf{y}\|^{2m-2d} d\mathbf{y} \int_\Omega \{Q_\alpha(\mathbf{x}, \mathbf{y}) D^\alpha \hat{g}_n(\mathbf{y})\}^2 d\mathbf{y} \right\}^{1/2}$$

$$\leq \max_{\mathbf{x}, \mathbf{y} \in \Omega} |Q_\alpha(\mathbf{x}, \mathbf{y})| \sum_{|\alpha|=m} \left\{ \int_\Omega \|\mathbf{x} - \mathbf{y}\|^{2m-2d} d\mathbf{y} \int_\Omega \{D^\alpha \hat{g}_n(\mathbf{y})\}^2 d\mathbf{y} \right\}^{1/2} \tag{B.3}$$

for $\mathbf{x} \in \Omega$.

Let $\rho$ be the radius of the smallest ball, centered at the origin, containing $\Omega$. Then, changing to the spherical coordinate system yields

$$\int_\Omega \|\mathbf{x} - \mathbf{y}\|^{2m-2d} d\mathbf{y} \leq \int_0^{2\pi} \int_0^{2\rho} r^{2(m-d)} r^{d-1} dr d\theta = 2\pi (2\rho)^{2(m-d/2)} \tag{B.4}$$

by the fact that $2m > d$.

By Assumption 7, (B.3), and (B.4), there exists a constant $M_2$ such that

$$\sup_{\mathbf{x} \in \Omega} |\hat{v}_n(\mathbf{x})| \leq M_2 \tag{B.5}$$

for $n$ sufficiently large a.s.

Step 3. We use (B.5) and the $\mathcal{P}_{m-1}$-unisolvency of $\{X_1, \cdots, X_n\}$ to show that $\sup_{\mathbf{x} \in \Omega} |\hat{u}_n(\mathbf{x})|$ is bounded uniformly on $n$ for $n$ sufficiently large a.s.

First, combining (B.2) and (B.5) yields

$$\frac{1}{n} \sum_{i=1}^{n} \hat{u}_n(X_i)^2 \leq \frac{2}{n} \sum_{i=1}^{n} \hat{g}_n(X_i)^2 + \frac{2}{n} \sum_{i=1}^{n} \hat{v}_n(X_i)^2 \leq 2M_1 + 2M_2^2 \tag{B.6}$$

for $n$ sufficiently large a.s.

Second, we consider the following matrix:

$$Z_n = \begin{pmatrix} p_1(X_1) & p_2(X_1) & \cdots & p_M(X_1) \\ \vdots & & & \vdots \\ p_1(X_n) & p_2(X_n) & \cdots & p_M(X_n) \end{pmatrix}.$$

By the $\mathcal{P}_{m-1}$-unisolvency of the $X_i$'s, the matrix $W \triangleq (1/n) Z_n^T Z_n$ is positive definite because $w^T W w = 0$ with $w^T = (w_1, \cdots, w_M)$ implies $\sum_{i=1}^{M} w_i p_i(X_j) = 0$ for all $1 \leq j \leq n$, which then implies $w_i = 0$ for all $1 \leq i \leq M$. Let $\lambda_1, \cdots, \lambda_M$ be the positive eigenvalues of $W$, and let $\rho_1, \cdots, \rho_M$ be the corresponding eigenvectors with $\|\rho_i\| = 1$ for $1 \leq i \leq M$. Let $\lambda_{min}$ be the minimum of $\lambda_1, \cdots, \lambda_M$.

We next prove that

$$\liminf_{n \to \infty} \lambda_{min} \geq \lambda_* > 0 \tag{B.7}$$

for some constant $\lambda_*$. Because $\rho_i^T W \rho_i = \lambda_i$ for $1 \leq i \leq M$, we have $\lambda_{min} \geq \min_{\|w\|=1} w^T W w$, where $w^T = (w_1, \cdots, w_M)$. So,

$$\lambda_{min} \geq \min_{w_1^2 + w_2^2 + \cdots + w_M^2 = 1} \frac{1}{n} \sum_{i=1}^{n} (w_1 p_1(X_i) + \cdots + w_M p_M(X_i))^2.$$

Note that $\Lambda = \{(w_1, \cdots, w_M) \in \mathbb{R}^M : w_1^2 + \cdots + w_M^2 = 1\}$ is a nonempty closed subset of $\mathbb{R}^M$, and

$$\frac{1}{n} \sum_{i=1}^{n} (w_1 p_1(X_i) + \cdots + w_M p_M(X_i))^2 \to \mathbb{E}(w_1 p_1(X) + \cdots + w_M p_M(X))^2$$

as $n \to \infty$ uniformly on $\Lambda$ because $\Lambda$ is bounded and the $X_i$'s are in $[a, b]^d$.

By Shapiro et al. [63, proposition 5.2 on p. 157],

$$\min_{w_1^2 + \cdots + w_M^2 = 1} \frac{1}{n} \sum_{i=1}^{n} (w_1 p_1(X_i) + \cdots + w_M p_M(X_i))^2 \to \min_{w_1^2 + \cdots + w_M^2 = 1} \mathbb{E}(w_1 p_1(X) + \cdots + w_M p_M(X))^2$$

a.s. as $n \to \infty$. Note that $\min_{w_1^2 + \cdots + w_M^2 = 1} \mathbb{E}(w_1 p_1(X) + \cdots + w_M p_M(X))^2 \neq 0$ because $\mathbb{E}(w_1 p_1(X) + \cdots + w_M p_M(X))^2 = 0$ implies $w_1 = \cdots = w_M = 0$. Hence,

$$\liminf_{n \to \infty} \lambda_{min} \geq \min_{w_1^2 + \cdots + w_M^2 = 1} \mathbb{E}(w_1 p_1(X) + \cdots + w_M p_M(X))^2 \triangleq \lambda_* > 0,$$

proving (B.7).

We finally prove that $\hat{u}_n$ is bounded on $\Omega$ uniformly on $n$ for $n$ sufficiently large. Let $\hat{u}_n$ be given by $\hat{u}_n = \hat{a}_{n,1} p_1 + \cdots + \hat{a}_{n,M} p_M$ for some $\hat{a}_{n,1}, \cdots, \hat{a}_{n,M}$. Let $\hat{a}_n^T \triangleq (\hat{a}_{n,1}, \cdots, \hat{a}_{n,M})$ and note that the Rayleigh–Ritz theorem implies

$$\frac{1}{n} \sum_{i=1}^{n} \hat{u}_n(X_i)^2 = \hat{a}_n^T W \hat{a}_n \geq \lambda_{min}(\hat{a}_{n,1}^2 + \cdots + \hat{a}_{n,M}^2) \geq \lambda_{min} \hat{a}_{n,i}^2 \tag{B.8}$$

for $1 \leq i \leq M$.

Combining (B.6), (B.7), and (B.8) yields

$$|\hat{a}_{n,i}| \leq \sqrt{(2M_1 + 2M_2^2)/(\lambda_*/2)} \tag{B.9}$$

for $1 \leq i \leq M$ and all $n$ sufficiently large a.s., and hence,

$$|\hat{u}_n(\mathbf{x})| = |\hat{a}_{n,1} p_1(\mathbf{x}) + \cdots + \hat{a}_{n,M} p_M(\mathbf{x})| \leq M \sqrt{(2M_1 + 2M_2^2)/(\lambda_*/2)} \max_{1 \leq i \leq M, \mathbf{x} \in \Omega} |p_i(\mathbf{x})| \triangleq M_3 \tag{B.10}$$

for all $\mathbf{x} \in \Omega$ and $n$ sufficiently large a.s., proving Step 3.

Combining (B.5) and (B.10) completes the proof of Lemma B.1. $\quad\square$

**Lemma B.2.** *Under Assumptions* 2, 3, 7, *and* 9, *there exists a constant* $\alpha_1$ *such that*

$$|\hat{g}_n(\mathbf{x}) - \hat{g}_n(\mathbf{y})| \leq \alpha_1 \|\mathbf{x} - \mathbf{y}\|^\delta$$

*for any* $\mathbf{x}, \mathbf{y} \in [a,b]^d$ *for* $n$ *sufficiently large a.s., where* $\delta = 1$ *when* $m \geq 2$ *and* $\delta = 1/2$ *when* $m = 1$.

**Proof of Lemma B.2.** We consider two cases separately: the case when $m = 1$ and the case when $m \geq 2$.

Case 1. $m = 1$. Because $2m > d$, it follows that $d = 1$. The fundamental theorem of calculus and the continuity of $\hat{g}_n$ imply

$$\hat{g}_n(x) - \hat{g}_n(y) = \int_y^x \hat{g}_n^{(1)}(t)dt,$$

where $\hat{g}_n^{(1)}$ is the weak derivative of $\hat{g}_n$ of the first order. By Assumption 7 and Hölder's inequality,

$$|\hat{g}_n(x) - \hat{g}_n(y)| \leq \left\{ \int_y^x \{\hat{g}_n^{(1)}(t)\}^2 dt \int_y^x dt \right\}^{1/2} \leq (c_B + 1)^{1/2} \|x - y\|^{1/2}$$

for $y \leq x$ and $n$ sufficiently large a.s., proving Lemma B.2.

Case 2. $m \geq 2$. Let $\Omega$ be a bounded open subset of $\mathbb{R}^d$ containing $[a,b]^d$.

We first prove that $D^\alpha \hat{g}_n(\mathbf{x})$ with $|\alpha| = 1$ is bounded uniformly on $n$ and $\mathbf{x}$ for $n$ sufficiently large a.s. By Oden and Reddy [54, equation 3.48 on p. 81] and Lemma B.1, $\limsup_{n\to\infty} \|\hat{g}_n\|_{W_n(\Omega)} < \infty$ a.s. Assumption 7 and Adams and Fournier [1, theorem 2 on p. 717] imply that there exists a constant $M_4$ such that

$$\max_{|\beta|=1} \int_\Omega \{D^\beta \hat{g}_n(\mathbf{x})\}^2 d\mathbf{x} \leq M_4 \quad \text{and} \quad \max_{|\beta|=2} \int_\Omega \{D^\beta \hat{g}_n(\mathbf{x})\}^2 d\mathbf{x} \leq M_4 \qquad (B.11)$$

for $n$ sufficiently large a.s.

When $\hat{g}_n$ is restricted to $\Omega$, $D^\alpha \hat{g}_n$ with $|\alpha| = 1$ belongs to $W_1(\Omega)$ for $n$ sufficiently large a.s. Applying the Sobolev integral identity (Oden and Reddy [54, theorem 3.6 on p. 68]) to $D^\alpha \hat{g}_n$ with $|\alpha| = 1$ yields

$$D^\alpha \hat{g}_n(\mathbf{x}) = \int_\Omega \zeta(\mathbf{y}) D^\alpha \hat{g}_n(\mathbf{y}) d\mathbf{y} + \int_\Omega \|\mathbf{x} - \mathbf{y}\|^{m-d} \sum_{|\beta|=2} Q_\beta(\mathbf{x}, \mathbf{y}) D^\beta \hat{g}_n(\mathbf{y}) d\mathbf{y}$$

for $\mathbf{x} \in \Omega$, where $\zeta(\mathbf{y})$ is a continuous bounded function of $\mathbf{y}$ and $Q_\beta(\mathbf{x}, \mathbf{y})$, $|\beta| = 2$, are bounded infinitely differentiable functions of $\mathbf{x}$ and $\mathbf{y}$. Hence, for $\mathbf{x} \in \Omega$ and $|\alpha| = 1$, combining Hölder's inequality, (B.4), and (B.11) yields

$$|D^\alpha \hat{g}_n(\mathbf{x})| \leq \left\{ \int_\Omega \zeta(\mathbf{y})^2 d\mathbf{y} \right\}^{1/2} \left\{ \int_\Omega D^\alpha \hat{g}_n(\mathbf{y})^2 d\mathbf{y} \right\}^{1/2}$$

$$+ \sum_{|\beta|=2} \left\{ \int_\Omega \|\mathbf{x} - \mathbf{y}\|^{2m-2d} d\mathbf{y} \right\}^{1/2} \left\{ \int_\Omega Q_\beta(\mathbf{x}, \mathbf{y})^2 \{D^\beta \hat{g}_n(\mathbf{y})\}^2 d\mathbf{y} \right\}^{1/2} \leq M_5 \qquad (B.12)$$

for some constant $M_5$ and $n$ sufficiently large a.s.

We next note that by the extended mean-value theorem, for any $\mathbf{x} = (x_1, \cdots, x_d)$ and $\mathbf{y} = (y_1, \cdots, y_d)$, satisfying $\mathbf{x}, \mathbf{x} + \mathbf{y} \in [a,b]^d$, we have

$$\hat{g}_n(\mathbf{x} + \mathbf{y}) - \hat{g}_n(\mathbf{x}) = \frac{\partial \hat{g}_n}{\partial x_1}(\mathbf{z}_1)y_1 + \cdots + \frac{\partial \hat{g}_n}{\partial x_d}(\mathbf{z}_d)y_d$$

for some $\mathbf{z} = \mathbf{x} + c\mathbf{y}$ with $c \in (0,1)$, and hence,

$$|\hat{g}_n(\mathbf{x} + \mathbf{y}) - \hat{g}_n(\mathbf{x})| \leq \{y_1^2 + \cdots + y_d^2\}^{1/2} \left\{ \frac{\partial \hat{g}_n}{\partial x_1}(\mathbf{z}_1)^2 + \cdots + \frac{\partial \hat{g}_n}{\partial x_d}(\mathbf{z}_d)^2 \right\}^{1/2} \leq M_6 \|\mathbf{y}\|.$$

This is for some constant $M_6$ and $n$ sufficiently large a.s.

Combining the two cases, we complete the proof of Lemma B.2. $\square$

**Lemma B.3.** *Under Assumptions* 2, 3, 7, *and* 9,

$$\frac{1}{n} \sum_{i=1}^n \varepsilon_i(\hat{g}_n(X_i) - f_*(X_i)) \to 0$$

*as* $n \to \infty$ *a.s.*

**Proof of Lemma B.3.** Let $\epsilon > 0$ be given. Let $\mathcal{A}_m$ be defined as in (B.1). Assumption 7, Lemma B.1, and Lemma B.2 imply that under Assumptions 2 and 3, any solution to Problem (B), restricted to $[a,b]^d$, belongs to $\mathcal{A}_m$ for $n$ sufficiently large a.s. By Kolmogorov and Tihomirov [39, theorem 13], there exist $l \triangleq l(\epsilon)$ and $h_1, \cdots, h_l$ in $\mathcal{A}_m$ such that for any $g \in \mathcal{A}_m$, there is $i \triangleq i(g) \in \{1, \cdots, l\}$

satisfying

$$\sup_{\mathbf{x}\in[a,b]^d} |g(\mathbf{x}) - h_l(\mathbf{x})| \le \epsilon. \tag{B.13}$$

For each $j \in \{1, \cdots, l\}$, we have

$$\frac{1}{n}\sum_{i=1}^{n} \varepsilon_i(\hat{g}_n(X_i) - f_*(X_i)) = \frac{1}{n}\sum_{i=1}^{n}\varepsilon_i(\hat{g}_n(X_i) - h_j(X_i)) + \frac{1}{n}\sum_{i=1}^{n}\varepsilon_i(h_j(X_i) - f_*(X_i))$$

$$\le \frac{1}{n}\sum_{i=1}^{n}|\varepsilon_i| \sup_{x\in[a,b]^d}|\hat{g}_n(x) - h_j(x)| + \max_{1\le j\le l}\frac{1}{n}\sum_{i=1}^{n}\varepsilon_i(h_j(X_i) - f_*(X_i)),$$

and hence,

$$\frac{1}{n}\sum_{i=1}^{n}\varepsilon_i(\hat{g}_n(X_i) - f_*(X_i))$$

$$\le \min_{1\le j\le l}\left(\sup_{x\in[a,b]^d}|\hat{g}_n(x) - h_j(x)|\right)\left(\frac{1}{n}\sum_{i=1}^{n}|\varepsilon_i|\right) + \max_{1\le j\le l}\frac{1}{n}\sum_{i=1}^{n}\varepsilon_i(h_j(X_i) - f_*(X_i))$$

$$\le \frac{\epsilon}{n}\sum_{i=1}^{n}|\varepsilon_i| + \max_{1\le j\le l}\frac{1}{n}\sum_{i=1}^{n}\varepsilon_i(h_j(X_i) - f_*(X_i)) \text{ by } (44)$$

$$\le \epsilon\mathbb{E}|\varepsilon| + \epsilon$$

for $n$ sufficiently large a.s. by the strong law of large numbers. $\quad\square$

**Lemma B.4.** *Under Assumptions* 2, 3, 7, *and* 9,

$$\frac{1}{n}\sum_{i=1}^{n}(\hat{g}_n(X_i) - f_*(X_i))^2 \to 0$$

*as* $n \to \infty$ *a.s.*

**Proof of Lemma B.4.** Note that

$$\frac{1}{n}\sum_{i=1}^{n}(\hat{g}_n(X_i) - f_*(X_i))^2 \le \frac{2}{n}\sum_{i=1}^{n}\varepsilon_i(\hat{g}_n(X_i) - f_*(X_i)) + S_n - \frac{1}{n}\sum_{i=1}^{n}\varepsilon_i^2.$$

Lemma B.4 follows from Assumption 9 and Lemma B.3. $\quad\square$

**Proof of Theorem 2.** We first use Lemma B.4 and the fact that $f_*$ is continuous over $[a,b]^d$ to conclude that $\hat{g}_n$ converges to $f_*$ uniformly over $[a,b]^d$ a.s.

Fix $\epsilon > 0$. We can find a finite number of sets $S_1, \cdots, S_l$ covering $[a,b]^d$, each having a diameter less than $\epsilon$ (i.e., $\sup\{\|\mathbf{x} - \mathbf{y}\| : \mathbf{x}, \mathbf{y} \in S_j\} \le \epsilon$) and having a nonempty interior. Because $f_* \in \mathcal{F}_m$ and hence, is continuous over $[a,b]^d$, Lemma B.2 implies that we can find a constant, say $c$, satisfying

$$|\hat{g}_n(\mathbf{x}) - \hat{g}_n(\mathbf{y})| \le c\|\mathbf{x} - \mathbf{y}\|^\delta \text{ and } |f_*(\mathbf{x}) - f_*(\mathbf{y})| \le c\|\mathbf{x} - \mathbf{y}\|^\delta$$

for $\mathbf{x}, \mathbf{y} \in [a,b]^d$ and $n$ sufficiently large a.s., where $\delta = 1$ when $m \ge 2$ and $\delta = 1/2$ when $m = 1$.

For each $\mathbf{x} \in S_j$ and $X_i \in S_j$,

$$|\hat{g}_n(\mathbf{x}) - f_*(\mathbf{x})| \le |\hat{g}_n(\mathbf{x}) - \hat{g}_n(X_i)| + |\hat{g}_n(X_i) - f_*(X_i)| + |f_*(X_i) - f_*(\mathbf{x})| \le c\epsilon^\delta + |\hat{g}_n(X_i) - f_*(X_i)| + c\epsilon^\delta,$$

so

$$\sup_{\mathbf{x}\in S_j}|\hat{g}_n(\mathbf{x}) - f_*(\mathbf{x})| \le 2c\epsilon^\delta + \frac{\sum_{i=1}^{n}|\hat{g}_n(X_i) - f_*(X_i)|I(X_i \in S_j)}{\sum_{i=1}^{n}I(X_i \in S_j)}$$

$$\le 2c\epsilon^\delta + \frac{1}{n}\sum_{i=1}^{n}|\hat{g}_n(X_i) - f_*(X_i)| \cdot \frac{n}{\sum_{i=1}^{n}I(X_i \in S_j))}$$

$$\le 2c\epsilon^\delta + \sqrt{\frac{1}{n}\sum_{i=1}^{n}(\hat{g}_n(X_i) - f_*(X_i))^2}\frac{n}{\sum_{i=1}^{n}I(X_i \in S_j))}.$$

By Lemma B.4 and Assumption 2, we conclude that

$$\limsup_{n\to\infty} \sup_{\mathbf{x}\in S_j} |\hat{g}_n(\mathbf{x}) - f_*(\mathbf{x})| \le 0$$

a.s. Because there are finitely many $S_j$'s, we conclude that

$$\sup_{\mathbf{x}\in[a,b]^d} |\hat{g}_n(\mathbf{x}) - f_*(\mathbf{x})| \to 0$$

a.s. as $n \to \infty$.

We next use Lemmas B.1, B.2, and B.4 to establish

$$\mathbb{E}(\hat{g}_n(X) - f_*(X))^2 \to 0 \tag{B.14}$$

as $n \to \infty$.

We need to use the truncated value defined as follows. For any $c \ge 0$ and $x \ge 0$, the truncated value of $x$, denoted by $T_c(x)$, is defined by

$$T_c(x) = \begin{cases} x, & \text{if } x \ge c \\ 0, & \text{otherwise.} \end{cases} \tag{B.15}$$

Note that $T_c(x^2) = \{T_{\sqrt{c}}(x)\}^2$ for $x \ge 0$, $T_c(x) \le T_c(y)$ for $0 \le x \le y$, and $T(x+y) \le T(x) + T(y)$ for $x, y \ge 0$. Together with the fact that $|x+y| \le |x| + |y|$ for $x, y \in \mathbb{R}$, we obtain

$$T_c(x+y)^2 \le T_c(x^2) + T_c(y^2) + 2T_{\sqrt{c}}|x| \cdot T_{\sqrt{c}}|y| \tag{B.16}$$

$$-T_c(x+y)^2 \le -T_c(x^2) - T_c(y^2) + 2T_{\sqrt{c}}|x| \cdot T_{\sqrt{c}}|y|. \tag{B.17}$$

We will prove that

$$\mathbb{E}T_c(\hat{g}_n(X) - f_*(X))^2 \to 0 \tag{B.18}$$

as $n \to \infty$ for each $c > 0$. Once (B.18) is proven, letting $c \uparrow \infty$ for each $n$ will prove (B.14). To show (B.18), let $\epsilon > 0$ and $c > 0$ be given. Let $\mathcal{A}_m$ be defined as in (B.1). Lemmas B.1 and B.2 imply that, under Assumption 2 and Assumption 3, any solution to Problem (B), restricted to $[a,b]^d$, belongs to $\mathcal{A}_m$ for $n$ sufficiently large a.s. By Kolmogorov and Tihomirov [39, theorem 13], there exist $l \triangleq l(\epsilon)$ and $h_1, \cdots, h_l$ in $\mathcal{A}_m$ such that for any $g \in \mathcal{A}_m$, there is $i \triangleq i(g) \in \{1, \cdots, l\}$ satisfying

$$\sup_{\mathbf{x}\in[a,b]^d} |g(\mathbf{x}) - h_i(\mathbf{x})| \le \epsilon. \tag{B.19}$$

For each $j \in \{1, \cdots, l\}$, we have

$$\mathbb{E}T_c(\hat{g}_n(X) - f_*(X))^2 - \frac{1}{n}\sum_{i=1}^n T_c(\hat{g}_n(X_i) - f_*(X_i))^2$$

$$= \mathbb{E}T_c(\hat{g}_n(X) - h_j(X) + h_j(X) - f_*(X))^2 - \frac{1}{n}\sum_{i=1}^n T_c(\hat{g}_n(X_i) - h_j(X_i) + h_j(X_i) - f_*(X_i))^2$$

$$\le \mathbb{E}T_c(\hat{g}_n(X) - h_j(X))^2 + \mathbb{E}(h_j(X) - f_*(X))^2 + 2\mathbb{E}T_{\sqrt{c}}|\hat{g}_n(X) - h_j(X)| \cdot T_{\sqrt{c}}|h_j(X) - f_*(X)|$$

$$\quad - \frac{1}{n}\sum_{i=1}^n T_c(\hat{g}_n(X_i) - h_j(X_i))^2 - \frac{1}{n}\sum_{i=1}^n T_c(h_j(X_i) - f_*(X_i))^2$$

$$\quad + \frac{2}{n}\sum_{i=1}^n T_{\sqrt{c}}|\hat{g}_n(X_i) - h_j(X_i)| \cdot T_{\sqrt{c}}|h_j(X_i) - f_*(X_i)| \text{ (B.16) and (B.17)}$$

$$\le \max_{1\le j\le l}\left\{ \mathbb{E}T_c(h_j(X) - f_*(X))^2 - \frac{1}{n}\sum_{i=1}^n T_c(h_j(X_i) - f_*(X_i))^2 \right\} + \mathbb{E}T_c(\hat{g}_n(X) - h_j(X))^2$$

$$\quad + 2\sqrt{\mathbb{E}T_c(\hat{g}_n(X) - h_j(X))^2}\sqrt{\mathbb{E}T_c(h_j(X) - f_*(X))^2} + \frac{1}{n}\sum_{i=1}^n T_c(\hat{g}_n(X_i) - h_j(X_i))^2$$

$$\quad + 2\sqrt{\frac{1}{n}\sum_{i=1}^n T_c(\hat{g}_n(X_i) - h_j(X_i))^2}\sqrt{\frac{1}{n}\sum_{i=1}^n T_c(h_j(X_i) - f_*(X_i))^2}$$

by the Cauchy–Schwarz inequality, so

$$\mathbb{E}T_c(\hat{g}_n(X) - f_*(X))^2 - \frac{1}{n}\sum_{i=1}^n T_c(\hat{g}_n(X_i) - f_*(X_i))^2$$

$$\leq \max_{1 \leq j \leq l}\left\{ \mathbb{E}T_c(h_j(X) - f_*(X))^2 - \frac{1}{n}\sum_{i=1}^n T_c(h_j(X_i) - f_*(X_i))^2 \right\} + \min_{1 \leq j \leq l}\mathbb{E}T_c(\hat{g}_n(X) - h_j(X))^2$$

$$+ \min_{1 \leq j \leq l}2\sqrt{\mathbb{E}T_c(\hat{g}_n(X) - h_j(X))^2}\sqrt{\mathbb{E}T_c(h_j(X) - f_*(X))^2} + \min_{1 \leq j \leq l}\frac{1}{n}\sum_{i=1}^n T_c(\hat{g}_n(X_i) - h_j(X_i))^2$$

$$+ \min_{1 \leq j \leq l}2\sqrt{\frac{1}{n}\sum_{i=1}^n T_c(\hat{g}_n(X_i) - h_j(X_i))^2}\sqrt{\frac{1}{n}\sum_{i=1}^n T_c(h_j(X_i) - f_*(X_i))^2}$$

$$\leq \epsilon + 2\epsilon^2 + 4\sqrt{c}\epsilon$$

a.s. for $n$ sufficiently large, proving

$$\mathbb{E}T_c(\hat{g}_n(X) - f_*(X))^2 - \frac{1}{n}\sum_{i=1}^n T_c(\hat{g}_n(X_i) - f_*(X_i))^2 \to 0$$

as $n \to \infty$. By the dominated convergence theorem, we have

$$\mathbb{E}T_c(\hat{g}_n(X) - f_*(X))^2 - \mathbb{E}\left[\frac{1}{n}\sum_{i=1}^n T_c(\hat{g}_n(X_i) - f_*(X_i))^2\right] \to 0$$

and

$$\mathbb{E}\left[\frac{1}{n}\sum_{i=1}^n T_c(\hat{g}_n(X_i) - f_*(X_i))^2\right] \to 0$$

as $n \to \infty$, and hence, (B.18) follows.

To prove the last part of Theorem 2, let $j \in \{1, \cdots, m-1\}$ be fixed. By Assumption 2(ii), for any $\alpha$ with $|\alpha| = j$,

$$\mathbb{E}(D^\alpha\hat{g}_n(X) - D^\alpha f_*(X))^2 \leq \tau_2 \sum_{|\beta| \leq j}\int_{[a,b]^d}(D^\beta\hat{g}_n(\mathbf{x}) - D^\beta f_*(\mathbf{x}))^2 d\mathbf{x}. \qquad (B.20)$$

By Adams and Fournier [1, third equality in theorem 2 on p. 717], there exists a constant $C$ such that the right-hand side of (B.20) is less than or equal to

$$\tau_2 C \sum_{|\beta| \leq m}\left\{\int_{[a,b]^d}(D^\beta\hat{g}_n(\mathbf{x}) - D^\beta f_*(\mathbf{x}))^2 d\mathbf{x}\right\}^{j/2m}\left\{\int_{[a,b]^d}(\hat{g}_n(\mathbf{x}) - f_*(\mathbf{x}))^2 d\mathbf{x}\right\}^{(m-j)/2m}$$

$$\leq (\tau_2/\tau_1)C \sum_{|\beta| \leq m}\left\{\int_{[a,b]^d}(D^\beta\hat{g}_n(\mathbf{x}) - D^\beta f_*(\mathbf{x}))^2 d\mathbf{x}\right\}^{j/2m}\{\mathbb{E}(\hat{g}_n(X) - f_*(X))^2\}^{(m-j)/2m}.$$

By Adams and Fournier [1, first equality of theorem 2 on p. 717] and the fact that $J(\hat{g}_n) \leq c_B + 1$ for $n$ sufficiently large a.s. and $\mathbb{E}(\hat{g}_n(X) - f_*(X))^2 \to 0$ as $n \to \infty$,

$$\sum_{|\beta| \leq m}\left\{\int_{[a,b]^d}(D^\beta\hat{g}_n(\mathbf{x}) - D^\beta f_*(\mathbf{x}))^2 d\mathbf{x}\right\}^{j/2m}$$

is bounded for $n$ sufficiently large a.s. Therefore, $\mathbb{E}(D^\alpha\hat{g}_n(X) - D^\alpha f_*(X))^2$ is bounded by a random variable, say $Z_n$, that converges to zero as $n \to \infty$ a.s. By using $T_c\mathbb{E}(D^\alpha\hat{g}_n(X) - D^\alpha f_*(X))^2 \leq T_c(Z_n)$ for any $c > 0$ and the dominated convergence theorem, we obtain $T_c\mathbb{E}(D^\alpha\hat{g}_n(X) - D^\alpha f_*(X))^2 \to 0$ as $n \to \infty$, and hence, it follows that $\mathbb{E}(D^\alpha\hat{g}_n(X) - D^\alpha f_*(X))^2 \to 0$ as $n \to \infty$. □

**Proof of Theorem 1.** We apply arguments similar to those in Theorem 2 and Lemmas B.1–B.4 to establish the following claims.

Step 1. By Assumption 6, $J(\hat{f}_n) \leq c_A + 1$ for $n$ sufficiently large a.s.

Step 2. Use arguments similar to those in the proof of Lemma B.1 to establish that Assumptions 2, 3, 6, and 8 imply that there exists a constant $\beta_0 \triangleq \beta_0(\Omega)$ such that $|\hat{f}_n(\mathbf{x})| \leq \beta_0$ for all $\mathbf{x} \in \Omega$ and $n$ sufficiently large a.s. for any bounded open subset $\Omega$ of $\mathbb{R}^d$ containing $[a,b]^d$.

Step 3. Use arguments similar to those in the proof of Lemma B.2 to establish that Assumptions 2, 3, 6, and 8 imply that there exists a constant $\beta_1$ such that $|\hat{f}_n(\mathbf{x}) - \hat{f}_n(\mathbf{y})| \leq \beta_1\|\mathbf{x} - \mathbf{y}\|^\delta$ for $\mathbf{x}, \mathbf{y} \in [a,b]^d$ and $n$ sufficiently large a.s., where $\delta = 1$ when $m \geq 2$ and $\delta = 1/2$ when $m = 1$.

Step 4. Use arguments similar to those in the proof of Lemma B.3 to establish that Assumptions 2, 3, 6, and 8 imply $(1/n)\sum_{i=1}^{n} \varepsilon_i(\hat{f}_n(X_i) - f_*(X_i)) \to 0$ as $n \to \infty$ a.s.

Step 5. Use arguments similar to those in the proof of Lemma B.4 to establish that Assumptions 2, 3, 6, and 8 imply $(1/n)\sum_{i=1}^{n}(\hat{f}_n(X_i) - f_*(X_i))^2 \to 0$ as $n \to \infty$ a.s.

Step 6. Use arguments similar to those in the proof of Theorem 2 to establish $\sup_{\mathbf{x} \in [a,b]^d} |\hat{f}_n(\mathbf{x}) - f_*(\mathbf{x})| \to 0$ as $n \to \infty$ a.s., $\mathbb{E}(\hat{f}_n(X) - f_*(X))^2 \to 0$ as $n \to \infty$, and $\mathbb{E}(D^\alpha \hat{f}_n(X) - D^\alpha f_*(X))^2 \to 0$ as $n \to \infty$ for any $\alpha$ satisfying $|\alpha| \le m - 1$. □

**Proof of Corollary 1.** It suffices to prove that Assumption 10 implies Assumption 6 and that Assumption 11 implies Assumption 8. Because $J(\hat{f}_n) \le U_n$ for all $n$, under Assumption 10, we have $\limsup_{n \to \infty} J(\hat{f}_n) \le \limsup_{n \to \infty} U_n < \infty$, so Assumption 6 holds. Note that Assumption 11 implies that $f_*$ is a feasible solution to Problem (A) for all $n$ sufficiently large a.s., so $E_n(\hat{f}_n) \le E_n(f_*)$, or equivalently, $\frac{1}{n}\sum_{i=1}^{n}(\hat{f}_n(X_i) - f_*(X_i))^2 \le \frac{2}{n}\sum_{i=1}^{n} \varepsilon_i(\hat{f}_n(X_i) - f_*(X_i))$. □

**Proof of Corollary 2.** It suffices to prove that Assumption 12 implies Assumption 9. Because $E_n(\hat{g}_n) \le S_n$,

$$\frac{1}{n}\sum_{i=1}^{n}(\hat{g}_n(X_i) - f_*(X_i))^2 \le \frac{2}{n}\sum_{i=1}^{n} \varepsilon_i(\hat{g}_n(X_i) - f_*(X_i)) + S_n - \frac{1}{n}\sum_{i=1}^{n} \varepsilon_i^2.$$

So, Assumption 12 implies Assumption 9. □

**Proof of Theorem 3.** Let $\tilde{f}_n$ be a solution to

$$\text{Minimize}_{f \in \mathcal{A}_m} E_n(f) \text{ subject to } J(f) \le U_n.$$

We note that

$$\left\{\frac{1}{n}\sum_{i=1}^{n}\left(\tilde{f}_n(X_i) - f_*(X_i)\right)^2\right\}^{1/2} = \mathcal{O}_p(n^{-m/(2m+d)}) \tag{B.21}$$

implies Theorem 3 because of Assumption 6 and Steps 2 and 3 in the proof of Theorem 1. Hence, the rest of the proof is devoted to proving (B.21).

By Assumption 13, we have

$$\frac{1}{n}\sum_{i=1}^{n}(\tilde{f}_n(X_i) - f_*(X_i))^2 \le \frac{2}{n}\sum_{i=1}^{n} \varepsilon_i(\tilde{f}_n(X_i) - f_*(X_i)). \tag{B.22}$$

Hence, if we can prove

$$\frac{1}{\sqrt{n}}\left|\sum_{i=1}^{n} \varepsilon_i(\tilde{f}_n(X_i) - f_*(X_i))\right| \Big/ \left\{\frac{1}{n}\sum_{i=1}^{n}(\tilde{f}_n(X_i) - f_*(X_i))^2\right\}^{(1/2)(1-d/2m)} = \mathcal{O}_p(1), \tag{B.23}$$

then the combination of (B.22) and (B.23) will prove (B.21).

To establish (B.23), we introduce the notion of $\epsilon$-covering sets and the covering number of a metric or pseudometric space $(T, d)$ as follows. For any $\epsilon > 0$, an $\epsilon$-covering set of $(T, d)$ is a class of functions in $T$ such that for any $f \in T$, there exists $h \in T$ satisfying $d(f, h) < \epsilon$. The covering number $N(\epsilon, T, d)$ of $(T, d)$ is the number of elements of a minimal covering set. In other words,

$$N(\epsilon, T, d) = \min\left\{L : \text{There exists } f_1, \cdots, f_L \text{ such that } T \subset \bigcup_{i=1}^{L} B(f_i, \epsilon)\right\},$$

where $B(f, \epsilon) = \{g \in T : d(f, g) \le \epsilon\}$ for $f \in T$.

We note that there exists a positive constant $\Gamma$ such that for each $\epsilon > 0$,

$$N_\infty(\epsilon) \triangleq \log(1 + N(\epsilon, \mathcal{A}_m, d_\infty)) \le \Gamma \epsilon^{-d/m}; \tag{B.24}$$

see, for example, Birman and Solomyak [5]. (B.24) implies

$$N_n(\epsilon) \triangleq \log(1 + N(\epsilon, \mathcal{A}_m, d_n)) \le \Gamma \epsilon^{-d/m}. \tag{B.25}$$

(B.25) together with Step 2 in the proof of Theorem 1 and Assumption 4 implies (B.23) by van de Geer [69, lemma 8.4], which completes the proof of Theorem 3. □

**Proof of Corollary 3.** In the proof of Corollary 2, we already proved that Assumption 10 implies Assumption 6. It suffices to prove that Assumption 11 implies Assumption 13. Under Assumption 11, $f_*$ is a feasible solution to Problem (A) for $n$ sufficiently large a.s., so $E_n(\hat{f}_n) \le E_n(f_*)$, or equivalently, $\frac{1}{n}\sum_{i=1}^{n}(\hat{f}_n(X_i) - f_*(X_i))^2 \le \frac{2}{n}\sum_{i=1}^{n} \varepsilon_i(\hat{f}_n(X_i) - f_*(X_i))$ for all $n$ sufficiently large a.s. □

**Proof of Theorem 4.** Let $L^2 = \{f : \mathbb{R}^d \to \mathbb{R} : \mathbb{E}(f(X)^2) < \infty\}$ be equipped with the seminorm $|f| \triangleq \{\mathbb{E}(f(X)^2)\}^{1/2}$ for $f \in L^2$. It should be noted that $L^2$ is a semi-Hilbert space and that $\mathcal{P}_{m-1}$ is a closed subspace of $L^2$. By Assumption 5(i), $f_* \in L^2$. By the projection theorem, there exists $f_\infty \in \mathcal{P}_{m-1}$ minimizing $\mathbb{E}(f(X) - f_*(X))^2$ over $\mathcal{P}_{m-1}$. Let $v^* \triangleq \mathbb{E}(f_\infty(X) - f_*(X))^2$. To prove that $v^* > 0$, suppose, on the contrary, that $\mathbb{E}(f_\infty(X) - f_*(X))^2 = 0$. By Assumption 5(iii) and Assumption 2, we reach $f_*(x) = f_\infty(x)$ for $x \in [a, b]^d$, which contradicts Assumption 5(ii).

To prove the second part of Theorem 4, let $f_P$ be defined as in Lemma 1. We will show that for $n$ sufficiently large,

$$0 < S_n \le E_n(f_P), \tag{B.26}$$

a.s., which implies that there exists a unique solution $\hat{g}_n$ to Problem (B) for $n$ sufficiently large a.s.

To show (B.26), we will first show that

$$E_n(f_P) \rightarrow \min_{(c_1, \cdots, c_M) \in \mathbb{R}^M} \mathbb{E}(c_1 p_1(X) + \cdots + c_M p_M(X) - f_*(X))^2 + \sigma^2 \tag{B.27}$$

as $n \rightarrow \infty$. To prove (B.27), we first note that $f_P$ is a solution to

$$\min_{(c_1, \cdots, c_M) \in \mathbb{R}^M} \frac{1}{n} \sum_{i=1}^{n} (c_1 p_1(X) + \cdots + c_M p_M(X) - f_*(X_i))^2$$

$$- \frac{2}{n} \sum_{i=1}^{n} \varepsilon_i (c_1 p_1(X) + \cdots + c_M p_M(X) - f_*(X_i)) + \frac{1}{n} \sum_{i=1}^{n} \varepsilon_i^2$$

and that

$$\frac{1}{n} \sum_{i=1}^{n} (c_1 p_1(X) + \cdots + c_M p_M(X) - f_*(X_i))^2$$

$$- \frac{2}{n} \sum_{i=1}^{n} \varepsilon_i (c_1 p_1(X) + \cdots + c_M p_M(X) - f_*(X_i)) + \frac{1}{n} \sum_{i=1}^{n} \varepsilon_i^2$$

$$\rightarrow \mathbb{E}(c_1 p_1(X) + \cdots + c_M p_M(X) - f_*(X))^2 + \sigma^2$$

as $n \rightarrow \infty$ a.s. by Assumption 3(ii), Assumption 3(iii), Assumption 5(i), and the strong law of large numbers. We then use Shapiro et al. [63, theorem 5.4 on p. 159] and the fact that

$$\mathbb{E}(c_1 p_1(X) + \cdots + c_M p_M(X) - f_*(X))^2 + \sigma^2 \tag{B.28}$$

is convex in $(c_1, \cdots, c_M)$. The only nontrivial condition of Shapiro et al. [63, theorem 5.4] is that the set of solutions to (B.28) is nonempty and bounded. In the first part of this proof, we showed that there exists a solution to (B.28). To prove that the set of solutions to (B.28) is bounded, we notice that $\mathbb{E}(c_1 p_1(X) + \cdots + c_M p_M(X))^2$ is positive definite because of Assumption 2 and the fact that $[a, b]^d$ contains a $\mathcal{P}_{m-1}$-unisolvent set, and hence, $\mathbb{E}(c_1 p_1(X) + \cdots + c_M p_M(X) - f_*(X))^2$ is coercive over $(c_1, \cdots, c_M) \in \mathbb{R}^M$. We then conclude that the set of solutions to (B.28) is bounded. Thus, the application of Shapiro et al. [63, theorem 5.4 on p. 159] yields (B.27). Because the minimizing value of (B.28) is $\sigma^2 + \nu^*$ and $\lim \sup_{n\rightarrow\infty} S_n < \sigma^2 + \nu^*$ a.s., we reach (B.26). □